\documentclass[twocolumn,10pt]{asme2ej}
\usepackage{comment}

\usepackage{setspace}
\usepackage{times}
\usepackage{graphicx}
\usepackage{url}
\usepackage{graphicx}
\usepackage{subcaption}
\usepackage{amsmath}
\usepackage{float}
\usepackage{multicol}
\usepackage{multirow}
\usepackage{datetime}
\newdateformat{customdate}{\ordinal{DAY} \monthname[\THEMONTH], \THEYEAR}

\usepackage{xcolor} 
\definecolor{mydarkgreen}{RGB}{0,200,0}
\definecolor{mydarkred}{RGB}{200,0,0}
\usepackage{hyperref}
\hypersetup{
    colorlinks=true,    
    citecolor=blue,     
    linkcolor=blue,     
    urlcolor=blue       
}
\usepackage{nomencl}

\usepackage{graphicx}
\usepackage{bbding}
\usepackage{tabularx,colortbl}
\usepackage{lscape}
\usepackage{adjustbox}
\usepackage{pdflscape}
\usepackage{setspace,lipsum}
\usepackage{caption}
\usepackage{titlesec}
\titlespacing*{\section}{0pt}{*0.5}{*0.5} 
\titlespacing*{\subsection}{0pt}{*0.5}{*0.5} 
\titlespacing*{\subsubsection}{0pt}{*0.5}{*0.5} 
\usepackage{enumitem}
\setlist[itemize]{itemsep=5pt, topsep=2pt} 
\setlist[enumerate]{noitemsep, topsep=1pt} 

\usepackage{soul}

\title{DrivAerNet: A Parametric Car Dataset for Data-Driven Aerodynamic Design and Prediction}

\title{DrivAerNet: A Parametric Car Dataset for Data-Driven Aerodynamic Design and Prediction} 

\author{Mohamed Elrefaie
\affiliation{
	Department of Mechanical Engineering\\
	Massachusetts Institute of Technology\\
	Cambridge, MA, 02139\\
    Email: mohamed.elrefaie@mit.edu
    }	
}
\author{Angela Dai
\affiliation{
	Department of Computer Science\\
	Technical University of Munich\\
	Garching, 85748 Germany\\
    Email: angela.dai@tum.de
    }	
}

\author{Faez Ahmed
\affiliation{
	Department of Mechanical Engineering\\
	Massachusetts Institute of Technology\\
	Cambridge, MA, 02139\\
    Email: faez@mit.edu
    }	
}

\begin{document}
\maketitle





\begin{abstract}
This study introduces DrivAerNet, a large-scale high-fidelity CFD dataset of 3D industry-standard car shapes, and RegDGCNN, a dynamic graph convolutional neural network model, both aimed at aerodynamic car design through machine learning. DrivAerNet, with its 4000 detailed 3D car meshes using 0.5 million surface mesh faces and comprehensive aerodynamic performance data comprising of full 3D pressure, velocity fields, and wall-shear stresses, addresses the critical need for extensive datasets to train deep learning models in engineering applications. 
It is 60\% larger than the previously available largest public dataset of cars, and is the only open-source dataset that also models wheels and underbody.
RegDGCNN leverages this large-scale dataset to provide high-precision drag estimates directly from 3D meshes, bypassing traditional limitations such as the need for 2D image rendering or Signed Distance Fields (SDF). By enabling fast drag estimation in seconds, RegDGCNN facilitates rapid aerodynamic assessments, offering a substantial leap towards integrating data-driven methods in automotive design. Together, DrivAerNet and RegDGCNN promise to accelerate the car design process and contribute to the development of more efficient cars.  To lay the groundwork for future innovations in the field, the dataset and code used in our study are publicly accessible at \url{https://github.com/Mohamedelrefaie/DrivAerNet}.
\end{abstract}

\begin{table*}[h!]
\renewcommand{\arraystretch}{1.5}
\setlength{\tabcolsep}{4.4pt}
\centering
{\footnotesize
\begin{tabular}{cccccccccccc}
\hline 
\multirow{2}{*}{Dataset} & \multirow{2}{*}{Size} & \multicolumn{4}{c}{Aerodynamics Data} & Wheels/Underbody & \multirow{2}{*}{Parametric}  & \multirow{2}{*}{\shortstack{Design \\ Parameters}} & \multirow{2}{*}{\shortstack{Simulation \\ Type}} & \multirow{2}{*}{\shortstack{Mesh \\ Resolution}} & \multirow{2}{*}{Open-source} \\
\cline{3-6}
& & $C_d$ & $C_l$ & $\textbf{u}$ & $p$ & Modeling & & & & & \\ 

\hline
Baque et al. 2018~\cite{baque2018geodesic} & 2,000 & \textcolor{mydarkgreen}{\CheckmarkBold} & \textcolor{mydarkred}{\XSolidBrush} & \textcolor{mydarkred}{\XSolidBrush} & \textcolor{mydarkred}{\XSolidBrush}  & \textcolor{mydarkred}{\XSolidBrush} &  \textcolor{mydarkgreen}{\CheckmarkBold} & 21 (3D) & RANS & - & \textcolor{mydarkred}{\XSolidBrush} \\
\hline
Umetani et al. 2018~\cite{Umetani2018} & 889 & \textcolor{mydarkgreen}{\CheckmarkBold} & \textcolor{mydarkred}{\XSolidBrush} & \textcolor{mydarkgreen}{\CheckmarkBold} & \textcolor{mydarkgreen}{\CheckmarkBold} & \textcolor{mydarkred}{\XSolidBrush} & \textcolor{mydarkred}{\XSolidBrush} & - & RANS & 700k & \textcolor{mydarkgreen}{\CheckmarkBold} \\
\hline
Gunpinar et al. 2019~\cite{Gunpinar2019} & 1,000 & \textcolor{mydarkgreen}{\CheckmarkBold} & \textcolor{mydarkred}{\XSolidBrush} & \textcolor{mydarkred}{\XSolidBrush} & \textcolor{mydarkred}{\XSolidBrush} & \textcolor{mydarkred}{\XSolidBrush} & \textcolor{mydarkgreen}{\CheckmarkBold} & 21 (2D) & RANS & 130k & \textcolor{mydarkred}{\XSolidBrush} \\
\hline
Remelli et al. 2020~\cite{remelli2020meshsdf} & 1,400 & \textcolor{mydarkred}{\XSolidBrush} & \textcolor{mydarkred}{\XSolidBrush} & \textcolor{mydarkred}{\XSolidBrush} & \textcolor{mydarkgreen}{\CheckmarkBold} & \textcolor{mydarkred}{\XSolidBrush} & \textcolor{mydarkred}{\XSolidBrush}  & - & - & - & \textcolor{mydarkred}{\XSolidBrush} \\
\hline
Jacob et al. 2021~\cite{Jacob2021} & 1,000 & \textcolor{mydarkgreen}{\CheckmarkBold} & \textcolor{mydarkgreen}{\CheckmarkBold} & \textcolor{mydarkgreen}{\CheckmarkBold} &  \textcolor{mydarkred}{\XSolidBrush} & \textcolor{mydarkgreen}{\CheckmarkBold} & \textcolor{mydarkgreen}{\CheckmarkBold} & 15 (3D) & LBM & 150M & \textcolor{mydarkred}{\XSolidBrush} \\
\hline
Usama et al. 2021~\cite{Usama2021} & 500 & \textcolor{mydarkgreen}{\CheckmarkBold} & \textcolor{mydarkred}{\XSolidBrush} & \textcolor{mydarkred}{\XSolidBrush} & \textcolor{mydarkred}{\XSolidBrush} & \textcolor{mydarkred}{\XSolidBrush} & \textcolor{mydarkgreen}{\CheckmarkBold} & 40 (2D) & - & - & \textcolor{mydarkred}{\XSolidBrush} \\
\hline
Rios et al. 2021~\cite{Rios2021} & 600 & \textcolor{mydarkgreen}{\CheckmarkBold} & \textcolor{mydarkgreen}{\CheckmarkBold} & \textcolor{mydarkred}{\XSolidBrush} & \textcolor{mydarkred}{\XSolidBrush} & \textcolor{mydarkred}{\XSolidBrush} & \textcolor{mydarkred}{\XSolidBrush} & - & RANS & 2M & \textcolor{mydarkred}{\XSolidBrush} \\
\hline
Song et al. 2023~\cite{song2023surrogate} & 2,474 & \textcolor{mydarkgreen}{\CheckmarkBold} & \textcolor{mydarkred}{\XSolidBrush} & \textcolor{mydarkred}{\XSolidBrush} & \textcolor{mydarkred}{\XSolidBrush} & \textcolor{mydarkred}{\XSolidBrush} & \textcolor{mydarkred}{\XSolidBrush} & - & URANS & 3M & \textcolor{mydarkgreen}{\CheckmarkBold} \\
\hline
\multirow{2}{*}{Li et al. 2023~\cite{li2023geometryinformed}} & 551 & \textcolor{mydarkgreen}{\CheckmarkBold} & \textcolor{mydarkred}{\XSolidBrush} & \textcolor{mydarkred}{\XSolidBrush} & \textcolor{mydarkgreen}{\CheckmarkBold} & \textcolor{mydarkred}{\XSolidBrush} & \textcolor{mydarkgreen}{\CheckmarkBold} & 6 (3D) & RANS & 7.2M & \textcolor{mydarkred}{\XSolidBrush} \\
\cline{2-12}
& 611 & \textcolor{mydarkgreen}{\CheckmarkBold} & \textcolor{mydarkred}{\XSolidBrush} & \textcolor{mydarkred}{\XSolidBrush} & \textcolor{mydarkgreen}{\CheckmarkBold} & \textcolor{mydarkred}{\XSolidBrush} & \textcolor{mydarkred}{\XSolidBrush} & - & RANS & 600-700k & \textcolor{mydarkred}{\XSolidBrush} \\
\hline
Trinh et al. 2024 \cite{trinh20243d} & 1,121 & \textcolor{mydarkred}{\XSolidBrush} & \textcolor{mydarkred}{\XSolidBrush} & \textcolor{mydarkgreen}{\CheckmarkBold} & \textcolor{mydarkgreen}{\CheckmarkBold} & \textcolor{mydarkred}{\XSolidBrush} & \textcolor{mydarkred}{\XSolidBrush} & - & - & - & \textcolor{mydarkred}{\XSolidBrush} \\
\hline
DrivAerNet (Ours) & 4,000 & \textcolor{mydarkgreen}{\CheckmarkBold} & \textcolor{mydarkgreen}{\CheckmarkBold} & \textcolor{mydarkgreen}{\CheckmarkBold} & \textcolor{mydarkgreen}{\CheckmarkBold} & \textcolor{mydarkgreen}{\CheckmarkBold} & \textcolor{mydarkgreen}{\CheckmarkBold} & 50 (3D) & RANS & 8-16M & \textcolor{mydarkgreen}{\CheckmarkBold} \\
\hline
\end{tabular}}

\caption{A comparative analysis of various aerodynamics datasets, highlighting key aspects such as the number of designs in the dataset (size), the inclusion of aerodynamic coefficients (drag coefficient $C_d$ and lift coefficient $C_l$), the inclusion of velocity ($\textbf{u}$) and pressure ($p$) fields, the modeling of wheels/underbody, the capacity for conducting parametric studies, the number of design parameters, simulation type, mesh resolution, and open-source availability. Note that RANS stands for Reynolds-Averaged Navier-Stokes, URANS for the unsteady case, and LBM stands for Lattice Boltzmann Method.}
\label{tab:datasets_comparison}
\vspace{-10pt}
\end{table*}

\section{Introduction}
Reducing fuel consumption and CO2 emissions through advanced aerodynamic design is crucial for the automobile industry. This can facilitate a faster transition towards electric cars, complementing the 2035 ban on internal combustion engine cars and aligning with the ambitious goal of achieving carbon neutrality by 2050 to combat global warming~\cite{mock2021pathways, brand2020road, martins2023assessing}. In aerodynamic design, navigating through intricate design choices involves a detailed examination of aerodynamic performance and design constraints, which is often slowed down by the time-consuming nature of high-fidelity Computational Fluid Dynamics (CFD) simulations and experimental wind tunnel tests. High-fidelity CFD simulations can take days to weeks per design~\cite{aultman2022evaluation}, while wind tunnel testing, despite its accuracy, is limited to examining only a handful of designs due to time and cost constraints. Data-driven approaches can alleviate this bottleneck by leveraging existing datasets to navigate through design and performance spaces, thereby speeding up the design exploration process and efficiently assessing aerodynamic designs.

Although recent advancements in data-driven approaches for aerodynamic design are promising, these advancements typically concentrate on simpler 2D cases~\cite{elrefaie2024surrogate, thuerey2020deep, ayman2023deep} or lower-fidelity CFD simulations~\cite{song2023surrogate, li2023geometryinformed, baque2018geodesic} and overlook the complexities inherent in real-world 3D designs and the challenges posed by high-fidelity CFD simulations. According to~\cite{heft2012experimental}, simplifying car designs by excluding components like wheels and mirrors, and not modeling the underbody, leads to a significant underestimation of aerodynamic drag. Taking these elements into account increased the drag by more than 1.4 times, highlighting the importance of detailed modeling for accurate aerodynamic analysis. 
Additionally, there is a dearth of publicly available high-fidelity car simulation datasets, potentially slowing down research in data-driven method development, as each researcher may need extensive computational resources to create their own data and test their methods on them. 

Responding to this challenge, our paper introduces DrivAerNet, a comprehensive dataset that features full 3D flow field information across 4000 high-fidelity car CFD simulations. It is made publicly available to serve as a benchmark for training deep learning models in aerodynamic assessment, generative design, and other machine learning applications.

To demonstrate the importance of large-scale datasets, we also develop a surrogate model for aerodynamic drag prediction based on Dynamic Graph Convolutional Neural Networks ~\cite{wang2019dynamic}. Our model, RegDGCNN, operates directly on extremely large 3D meshes, eliminating the necessity for 2D image rendering or Signed Distance Fields (SDF) generation.
RegDGCNN's ability to swiftly identify aerodynamic improvements opens new avenues for creating more efficient cars by streamlining the evaluation of design adjustments. It marks a significant step towards optimizing car designs more efficiently.

Overall, the contributions of this paper are:
\begin{itemize}
    \item The release of DrivAerNet, an extensive high-fidelity dataset featuring 4000 car designs, complete with detailed 3D models with 0.5 million surface mesh faces each, full 3D flow fields, and aerodynamic performance coefficients. The dataset is \textbf{60\%} larger than the previously available largest public dataset of cars, and is the only open-source dataset that also models wheels and underbody, allowing accurate estimation of drag.
     \item The introduction of a surrogate model, named RegDGCNN, based on Dynamic Graph Convolutional Neural Networks for prediction of aerodynamic drag.
    RegDGCNN outperforms state-of-the-art attention-based models~\cite{Arechiga2023, song2023surrogate} for drag prediction by \textbf{3.57$\%$} on the ShapeNet benchmark dataset while using \textbf{1000$\times$} fewer parameters, and achieves an $R^2$ score of \textbf{0.9} on the DrivAerNet dataset.

\end{itemize}

In addition, the large size of our dataset is also justified by our analysis in Section \ref{sec:TrainingSize}, which reveals that expanding the training dataset from 560 to 2800 car designs from DrivAerNet resulted in a \textbf{75$\%$} decrease in error, illustrating the direct correlation between dataset size and model performance. A similar trend is observed with the dataset from~\cite{song2023surrogate}, where enlarging the number of training samples from 1270 to 6352 entries yielded a \textbf{56$\%$} error reduction, further validating our model's efficacy and the inherent value of large datasets in driving advancements in surrogate modeling.

The structure of our paper is as follows: 
Section~\ref{sec:RelatedWork} provides an overview of related work. 
Section~\ref{sec:TurbulenceModeling} introduces the fundamentals of turbulence modeling, beginning with the Navier-Stokes equations, Reynolds averaging, and the \(k-\omega\) SST turbulence model, which is used for generating our dataset. 
Section~\ref{sec:DrivAerNetDataset} presents the DrivAerNet dataset, detailing the numerical simulation methods, CFD results, geometric feasibility analysis, and dataset characteristics. 
In Section~\ref{sec:RegDGCNN}, we introduce our RegDGCNN approach using the Dynamic Graph Convolutional Neural Network for regression tasks. 
Section~\ref{sec:SurrogateModeling} examines the application of the RegDGCNN model for surrogate modeling of aerodynamic drag, comparing its performance on the DrivAerNet and ShapeNet datasets and underscoring the benefits of scaling the training dataset. 
The paper highlights the limitations of our study and suggests avenues for future research, followed by a conclusion that summarizes the results and key findings.

\section{Related Work}
\label{sec:RelatedWork}
This section starts with an overview of aerodynamics datasets and then transitions to discussing recent progress in 3D learning for aerodynamics.
\subsection{Aerodynamics Datasets}
Data-driven aerodynamic design is a methodology that leverages computational models and machine learning algorithms to optimize car shapes based on large volumes of aerodynamic performance data, aiming to improve efficiency.
A common type of data-driven aerodynamic design is surrogate modeling, which uses simplified models to approximate the behavior of complex aerodynamic phenomena, enabling faster simulations and iterations in the design process. It is particularly useful for preliminary design phases where quick evaluations are necessary.
However, a significant portion of existing research on data-driven aerodynamic design is concentrated on simplified 2D scenarios such as airfoils/2D geometries~\cite{bonnet2022airfrans, thuerey2020deep, elrefaie2024surrogate, Usama2021,kashefi2022physics, ayman2023deep,  Gunpinar2019} or simplified 3D models~\cite{baque2018geodesic, remelli2020meshsdf, li2023geometryinformed,Umetani2018,Rios2021,song2023surrogate,trinh20243d}. While these studies are instrumental in understanding the fundamental physics, they often fall short in terms of applicability to complex 3D real-world problems. This gap is further widened by the absence of proprietary data from the industry, posing challenges in replicating and validating research findings. Moreover, the absence of a standardized benchmark dataset in the field hampers the ability to consistently evaluate and compare the efficacy of various machine learning methodologies. This stands in contrast to fields such as image processing or 3D shape analysis, where benchmark datasets like ImageNet~\cite{deng2009imagenet} or ShapeNet~\cite{chang2015shapenet} have accelerated significant advancements in deep learning by providing a common ground for method comparison.
Although benchmarks for CFD simulations and turbulence modeling~\cite{ashton2023summary, ashton2022overview,aultman2022evaluation} do exist, their limited size often renders them inadequate for the demands of deep learning techniques, underscoring the necessity for a broader and more comprehensive dataset. 

Table ~\ref{tab:datasets_comparison} provides an overview of existing datasets in the literature for data-driven aerodynamic design. It compares their size, the inclusion of aerodynamic information such as drag coefficient ($C_d$), lift coefficient ($C_l$), velocity field (\(\textbf{u}\)), pressure (\(p\)), whether they are parametric, the number of design parameters, and their availability as open-source. In addition, we consider the modeling of rotating wheels and underbody for evaluation. The table also introduces details on the type of simulation employed, distinguishing between Reynolds-Averaged Navier-Stokes (RANS), Unsteady Reynolds-Averaged Navier-Stokes (URANS), and Lattice Boltzmann Method (LBM), as well as the mesh resolution used in each dataset. This comprehensive comparison highlights the strengths and limitations of each dataset in terms of aerodynamic performance and computational resources. Our dataset stands out with the largest size of 4,000 samples, comprehensive aerodynamic information, parametric details, modeling of the wheels and underbody, and open-source accessibility. 

Below, we discuss recent advancements in 3D learning for aerodynamics.

\subsection{Advancements in 3D Learning for Car Aerodynamics}
The study from \cite{trinh20243d} presented a super-resolution model aimed at refining the estimated, yet coarsely resolved, flow fields around cars from deep learning predictions to a higher resolution, crucial for aerodynamic car design. By incorporating a residual-in-residual dense block (RRDB) within the generator structure and employing a relativistic discriminator for enhanced detail capture, coupled with a novel distance-weighted loss and physics-informed loss to ensure physical accuracy, the methodology demonstrated a marked improvement in flow field enhancement around cars, outperforming previous approaches in this domain. 
Jacob et al.~\cite{Jacob2021} demonstrated that deep learning, particularly a modified U-Net architecture using SDF, can accurately predict drag coefficients for a specific car design without the need for explicit parameterization.  Kontou et al.~\cite{kontou2023dnn} introduced a DNN-based surrogate model for turbulence and transition closure in RANS equations, optimized using a Metamodel-Assisted Evolutionary Algorithm (MAEA). The DNN replaces the numerical solution of the turbulence model's partial differential equations (PDEs), significantly reducing computational time in CFD-based shape optimization tasks. The RANS-DNN model demonstrated notable performance improvements in 3D aerodynamic cases, including the DrivAer car model~\cite{heft2012introduction}, achieving efficient and accurate predictions of pressure fields and optimized geometries.
 
In recent developments in computational design, many studies have leveraged the ShapeNet~\cite{chang2015shapenet} dataset for shape optimization and surrogate modeling in aerodynamics. ShapeNet is a dataset consisting of millions of 3D models spanning 55 common object categories, designed to support research in computer vision, robotics, and geometric deep learning. Using ShapeNet, Remelli et al.~\cite{remelli2020meshsdf} enhanced shape optimization by employing MeshSDF, offering a more flexible alternative to traditional hand-crafted parameterizations. Song et al. ~\cite{song2023surrogate} introduced a novel 2D representation for 3D shapes, coupled with a surrogate model for aerodynamic drag prediction, showcasing the potential for AI-driven design optimizations. Rios et al. ~\cite{Rios2021} compared various design representation methods, including PCA, kernel-PCA, and a 3D point cloud autoencoder, highlighting the autoencoder's capability for localized shape modifications in aerodynamic optimizations. The geometry-informed neural operator (GINO)~\cite{li2023geometryinformed} was introduced, leveraging SDF, point clouds, and neural operators for efficient large-scale simulations, achieving significant speed-ups and error rate reductions in predicting surface fields (e.g., pressure and wall shear stress) and drag coefficients of different car geometries.  Deng et al.~\cite{denggeometry} proposed 3D Geometry-guided conditional adaptation (3D-GeoCA), a model-agnostic framework that enhances surrogate models for solving PDEs on arbitrary 3D geometries. Applied to the ShapeNet Car~\cite{Umetani2018} and Ahmed-Body~\cite{li2023geometryinformed} datasets, 3D-GeoCA significantly reduces the L-2 error in large-scale Reynolds-Averaged Navier-Stokes simulations, offering a robust solution for handling complex geometries in aerodynamic modeling.

Building on these developments, transformers have recently been applied to PDEs with the introduction of Transolver~\cite{wu2024transolver}, a novel approach utilizing Physics-Attention to group mesh points based on similar physical states. This method enables more efficient and accurate modeling of complex geometries, achieving state-of-the-art performance on the ShapeNet dataset~\cite{Umetani2018}.

\paragraph{Limitations of Existing Studies and the Impact of Comprehensive Modeling:}
Despite the novel methodologies employed in these studies,  they faced limitations stemming from the inherent drawbacks of the ShapeNet dataset, such as lower mesh resolution, lack of watertight geometries, small dataset size, and oversimplifications like modeling cars as single-bodied entities without detailed considerations for components like wheels, underbody, and side mirrors, which can significantly impact real-world aerodynamic performance. According to Heft et al.~\cite{heft2012experimental}, including these details in the DrivAer fastback model increased the drag value from 0.115 to 0.278 in CFD simulations and from 0.125 to 0.275 in wind tunnel experiments. This represents a substantial increase in drag by approximately 142$\%$ and 120$\%$ respectively, underscoring the critical role of comprehensive modeling in achieving accurate aerodynamic assessments. Another common hurdle in both surrogate modeling and design optimization is the scarcity of data, which complicates efforts to replicate results or benchmark various models and approaches. Addressing this challenge, our contribution introduces DrivAerNet, a comprehensive benchmark dataset tailored for data-driven aerodynamic design, aiming to facilitate comparison and validation of future methodologies.

\section{Turbulence Modeling}
\label{sec:TurbulenceModeling}
In this section, we start with the basics of turbulence modeling, beginning with the Navier-Stokes equations. Next, we discuss Reynolds averaging, a key method in turbulence modeling. We then introduce the \( k-\omega \) SST turbulence model, chosen for its effectiveness in accurately simulating a wide range of conditions, which makes it ideal for creating our dataset. The following derivations are summarized from \cite{wilcox1998turbulence, menter2003ten,  spalding1974numerical}.

\subsection{Reynolds-Averaged Navier–Stokes Equations}
The Navier-Stokes equations model fluid flow across various applications, from laminar to turbulent motion in science and engineering. Based on Newton's second law, they describe fluid dynamics but are highly nonlinear and coupled, making analytical solutions infeasible except in simple cases. Numerical methods and CFD techniques are typically required for solving these second-order PDEs. The component-wise form of the Navier-Stokes equations for incompressible flow in Cartesian coordinates is essential for practical applications, allowing for the detailed modeling of fluid behavior in three-dimensional space.
\begin{align}
& \quad \quad \quad \quad \frac{\partial u_1}{\partial x_1} + \frac{\partial u_2}{\partial x_2} + \frac{\partial u_3}{\partial x_3} = 0 \\
&\rho\left(\frac{\partial u_1}{\partial t} + u_1 \frac{\partial u_1}{\partial x_1} + u_2 \frac{\partial u_1}{\partial x_2} + u_3 \frac{\partial u_1}{\partial x_3}\right) = \notag \\
&\quad -\frac{\partial p}{\partial x_1} + \mu\left(\frac{\partial^2 u_1}{\partial x_1^2} + \frac{\partial^2 u_1}{\partial x_2^2} + \frac{\partial^2 u_1}{\partial x_3^2}\right) + \rho f_1 \\
&\rho\left(\frac{\partial u_2}{\partial t} + u_1 \frac{\partial u_2}{\partial x_1} + u_2 \frac{\partial u_2}{\partial x_2} + u_3 \frac{\partial u_2}{\partial x_3}\right) = \notag \\
&\quad -\frac{\partial p}{\partial x_2} + \mu\left(\frac{\partial^2 u_2}{\partial x_1^2} + \frac{\partial^2 u_2}{\partial x_2^2} + \frac{\partial^2 u_2}{\partial x_3^2}\right) + \rho f_2 \\
&\rho\left(\frac{\partial u_3}{\partial t} + u_1 \frac{\partial u_3}{\partial x_1} + u_2 \frac{\partial u_3}{\partial x_2} + u_3 \frac{\partial u_3}{\partial x_3}\right) = \notag \\
&\quad -\frac{\partial p}{\partial x_3} + \mu\left(\frac{\partial^2 u_3}{\partial x_1^2} + \frac{\partial^2 u_3}{\partial x_2^2} + \frac{\partial^2 u_3}{\partial x_3^2}\right) + \rho f_3
\end{align}

Here, \( u_1, u_2, u_3 \) represent the velocity components in the \( x_1, x_2, x_3 \) directions, respectively, while \( p \) is the pressure, \( \rho \) is the fluid density, \( \mu \) is the dynamic viscosity, and \( f_1, f_2, f_3 \) are the body force components acting in the respective directions. The terms account for the fluid's inertia, pressure gradient, viscous diffusion, and external forces. 
The mass conservation and impulse conservation can be written in Einstein summation notation as:
\vspace{-10pt}
\begin{equation}
\begin{gathered}
\frac{\partial u_{i}}{\partial x_{i}}=0 \\
\frac{\partial u_{i}}{\partial t}+\frac{\partial u_{i} u_{j}}{\partial x_{j}}+\frac{1}{\rho} \frac{\partial p}{\partial x_{i}}-v \frac{\partial^{2} u_{i}}{\partial x_{k}^{2}}=0
\end{gathered}    
\end{equation}

\paragraph{Reynolds-Averaging}  
Reynolds averaging separates the fluctuating and mean components of a turbulent flow, simplifying analysis by focusing on the mean flow properties. It is widely used in turbulence modeling for velocity, pressure, and other flow quantities. The incompressible Navier-Stokes equations for the mean values of the velocity field under Reynolds averaging are given by:
\vspace{-5pt}
\begin{equation}
\frac{\partial\left\langle u_{i}\right\rangle}{\partial t} + \frac{\partial\left\langle u_{i}\right\rangle\left\langle u_{j}\right\rangle}{\partial x_{j}} + \frac{1}{\rho} \frac{\partial\langle p\rangle}{\partial x_{i}} - v \frac{\partial^{2}\left\langle u_{i}\right\rangle}{\partial x_{k}^{2}} = -\frac{\partial\left\langle u_{i}^{\prime} u_{j}^{\prime}\right\rangle}{\partial x_{j}}
\end{equation}
This equation represents the conservation of momentum for the mean velocity field \( \left\langle u_{i} \right\rangle \), including contributions from the Reynolds stresses \( \tau_{i j} \), which are defined as the average of the product of velocity fluctuations \( u_i^{\prime} \) and \( u_j^{\prime} \).
\vspace{-10pt}

\begin{equation}
\tau_{ij} = \left\langle u_{i}^{\prime} u_{j}^{\prime}\right\rangle
\end{equation}

The Reynolds stress tensor \( \tau_{i j} \) plays a critical role in these equations, capturing the effects of turbulence on the mean flow.

\begin{equation}
\frac{\partial\left\langle u_{i}\right\rangle}{\partial x_{i}} = 0
\end{equation}
This equation represents the incompressibility condition, ensuring that the divergence of the mean velocity field is zero.

\paragraph{$k-\omega$ \text{Shear Stress Transport (SST)}}

The $k-\omega$ Shear Stress Transport (SST) model is a two-equation turbulence model that focuses on turbulence kinetic energy ($k$) and specific dissipation rate ($\omega$). It is designed to address the limitations of the standard $k-\omega$ model, particularly its sensitivity to freestream values of $k$ and $\omega$. The SST model is particularly effective at accurately depicting flow separation phenomena, making it a popular choice in aerodynamic and hydrodynamic simulations.

\begin{figure*}[h!]
    \centering
    \includegraphics[width=0.8\textwidth]{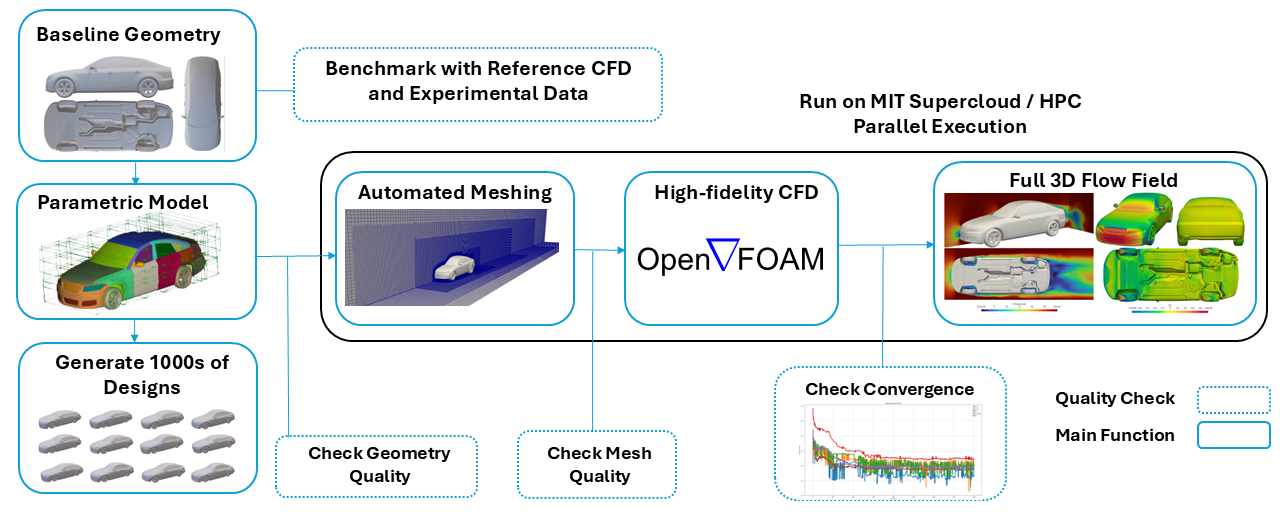}
\caption[Workflow for generating the car designs and validating the CFD simulations.]{The workflow begins with a baseline model for which existing CFD and experimental data are available, allowing for the benchmarking and validation of the simulations. At this stage, a balance between computational cost and accuracy is established. Next, a parametric model is created, enabling the generation of thousands of car designs. Automated meshing is then applied to these designs before high-fidelity CFD simulations are conducted using OpenFOAM. The resulting CFD data is stored for further analysis. Multiple quality checks are integrated throughout the process to ensure the accuracy and convergence of the simulations.}
    \label{fig:DrivAerNet_pipeline}
\end{figure*}

\paragraph{Model Equations}

The turbulence specific dissipation rate equation is given by:

\begin{align}
    \frac{D}{D t}(\rho \omega) = & \ \nabla \cdot\left(\rho D_\omega \nabla \omega\right) + \frac{\rho \gamma G}{\nu} - \frac{2}{3} \rho \gamma \omega (\nabla \cdot \mathbf{u}) \nonumber \\
    & - \rho \beta \omega^2 - \rho \left(F_1 - 1\right) C D_{k\omega} + S_\omega,
\end{align}

\noindent where:
\begin{itemize}
  \item \( D_\omega \) is the diffusivity of \( \omega \).
  \item \( G \) represents the generation of turbulence kinetic energy due to mean velocity gradients.
  \item \( \gamma \) and \( \beta \) are model constants.
  \item \( S_\omega \) is the source term for \( \omega \).
\end{itemize}
The equation for turbulence kinetic energy is given by:

$$
\frac{D}{D t}(\rho k) = \nabla \cdot\left(\rho D_k \nabla k\right) + \rho G - \frac{2}{3} \rho k (\nabla \cdot \mathbf{u}) - \rho \beta^* \omega k + S_k,
$$

\noindent where:
\begin{itemize}
  \item \( D_k \) is the diffusivity of \( k \).
  \item \( \beta^* \) is a model constant.
  \item \( S_k \) is the source term for \( k \).
\end{itemize}
The turbulence viscosity \( \nu_t \) is obtained using the following relation:

$$
\nu_t = a_1 \frac{k}{\max \left(a_1 \omega, b_1 F_{23} \mathbf{S}\right)},
$$

\noindent where:
\begin{itemize}
  \item \( a_1 \) and \( b_1 \) are model constants.
  \item \( F_{23} \) is a blending function that ensures the model transitions smoothly between the $k-\omega$ model behavior in the near-wall region and the $k-\varepsilon$ model behavior in the free stream.
  \item \( \mathbf{S} \) represents the magnitude of the mean rate-of-strain tensor.
\end{itemize}
The SST model integrates distinct advantages in fluid dynamics simulations by employing a blending function \( F_{23} \) to effectively merge the strengths of the \( k-\omega \) model, which excels in near-wall regions, with the \( k-\varepsilon \) model, preferred for free-stream and far-field applications. This model stands out for its precise predictions of flow separation under adverse pressure gradients, enhancing its utility across various engineering disciplines including aerodynamics and hydrodynamics~\cite{wilcox1998turbulence, menter2003ten}. 

For the DrivAerNet dataset generation, the $k-\omega$ SST turbulence model will be used to conduct the high-fidelity CFD simulations.

\section{DrivAerNet Dataset}
\label{sec:DrivAerNetDataset}

In this section, the workflow for generating the DrivAerNet dataset is presented in detail, as illustrated in Figure~\ref{fig:DrivAerNet_pipeline}. The process begins with the selection of a baseline car model, chosen because it has comprehensive CFD and experimental data available. This baseline is critical for benchmarking and validating the accuracy of the simulations that follow. At this stage, a careful balance is struck between computational efficiency and simulation accuracy. Following this, a detailed parametric model is developed, enabling the generation of thousands of unique car designs. Automated meshing is then applied to these designs, preparing them for high-fidelity CFD simulations using OpenFOAM~\cite{OpenFOAMv11}. The resulting simulation data is systematically stored for subsequent analysis. Throughout the entire workflow, multiple quality control measures are integrated to ensure that the simulations not only converge but also maintain high accuracy.

\paragraph{Introduction to DrivAerNet Dataset and Model Background:}

The goal is to create a large-scale dataset of 3D car shapes with high-quality mesh resolutions, watertight geometries, and diverse configurations that resemble real-world car designs. The ShapeNet~\cite{chang2015shapenet} dataset offers a wide range of car shapes; however, the mesh resolution and non-watertightness make it unsuitable for large-scale engineering designs or CFD datasets. Other generic car models, like the Ahmed body~\cite{ahmedbody} or the Windsor body~\cite{windsorbody}, are very simple and do not represent actual cars with realistic shapes.

\begin{figure*}[h!]
    \centering
    \includegraphics[width=0.8\textwidth]{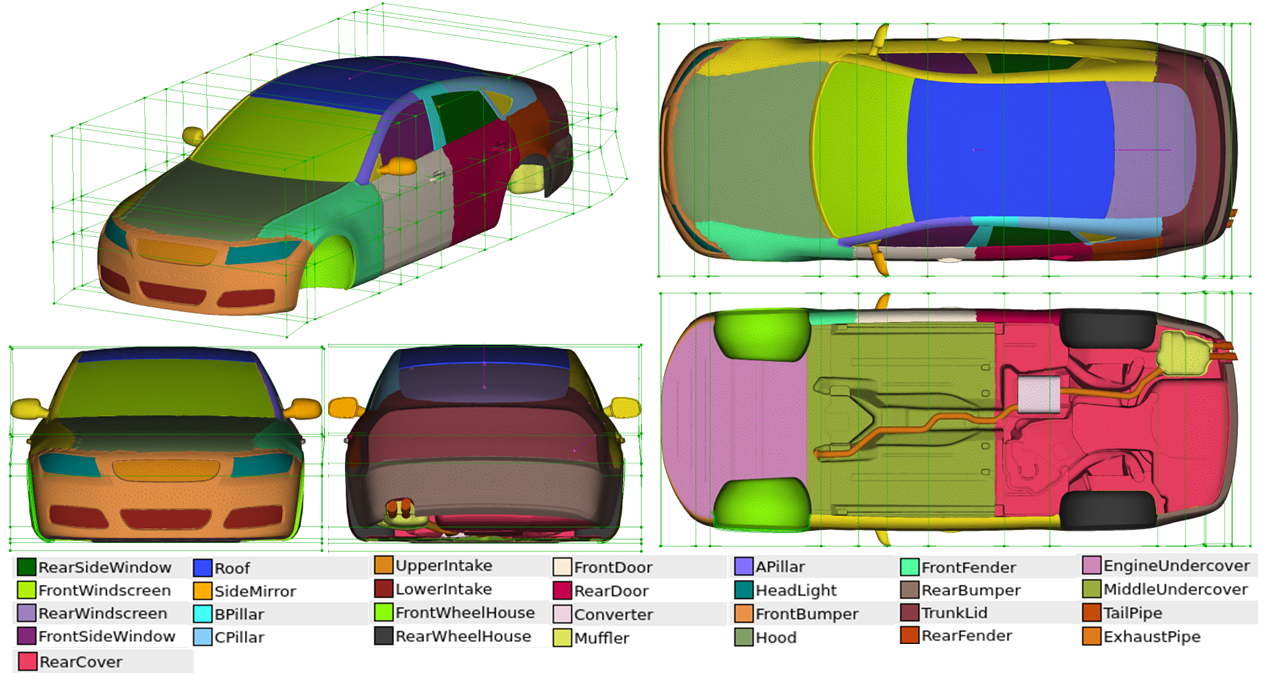}
    \caption{The parametric DrivAer model is depicted with morphing boxes applied for geometry transformation in ANSA® software using a total of 50 geometric parameters and 32 morphable entities. The morphing boxes are color-coded to highlight the areas susceptible to parametric modifications, facilitating the creation of 'DrivAerNet' dataset. Utilizing this morphing technique, we generated 4,000 unique car designs.}
    \label{fig:DrivAer++Parametric}
    \vspace{-10pt} 

\end{figure*}

DrivAer model~\cite{heft2012introduction, drivAergeometry} is a well-established conventional car reference model developed by researchers at the Technical University of Munich (TUM). It is a combination of the BMW series 3 and Audi A4 car designs in order to be representative of most conventional cars. The DrivAer model was developed to bridge the gap between open-source oversimplified models like the Ahmed~\cite{ahmedbody} and SAE~\cite{cogotti1998parametric} bodies and the complex designs of the manufacturing companies, which are not publicly available. 
To accurately assess real-world aerodynamic designs, we selected the fastback configuration with a detailed underbody, wheels, and mirrors (FDwWwM) as our baseline model, as shown in Figure~\ref{fig:DrivAer++Parametric}. This choice of the FDwWwM model was driven by the substantial impact of wheels, mirrors, and underbody geometry on aerodynamic drag, a conclusion supported by the findings in~\cite{heft2012experimental}. Specifically, the detailed underbody geometry adds 32-34 counts~\footnote{In aerodynamic evaluations, ``drag counts'' are used as a unit of measurement to denote small changes in the drag coefficient (\(C_d\)). One drag count is defined as a \(0.0001\) increment in \(C_d\).} to the drag, the inclusion of mirrors introduces an additional 14-16 counts, and the presence of wheels elevates the total drag coefficient by 102 counts.

\paragraph{Selecting the Baseline Parametric Model:}

In order to create a comprehensive dataset for training deep learning models for surrogate modeling and design optimization, we first created a parametric model of the DrivAer model. This approach was necessitated by the limitations of the original model, which was provided as a single, non-parametric STL mesh. To adequately capture the geometric variations and design modifications relevant to real-world automotive challenges, we developed a version of the DrivAer model that is defined by 50 geometric parameters and includes 32 morphable entities (see Figure~\ref{fig:DrivAer++Parametric}) using the commercial software ANSA®.
This parametric model allows for a more detailed exploration of the design space, facilitating the generation of 4,000 unique design variants through the application of the Optimal Latin Hypercube sampling method, specifically employing the Enhanced Stochastic Evolutionary Algorithm (ESE) as outlined by~\cite{damblin2013numerical}.  
\begin{figure}[!h]
    \centering
    \includegraphics[width=0.8\columnwidth]{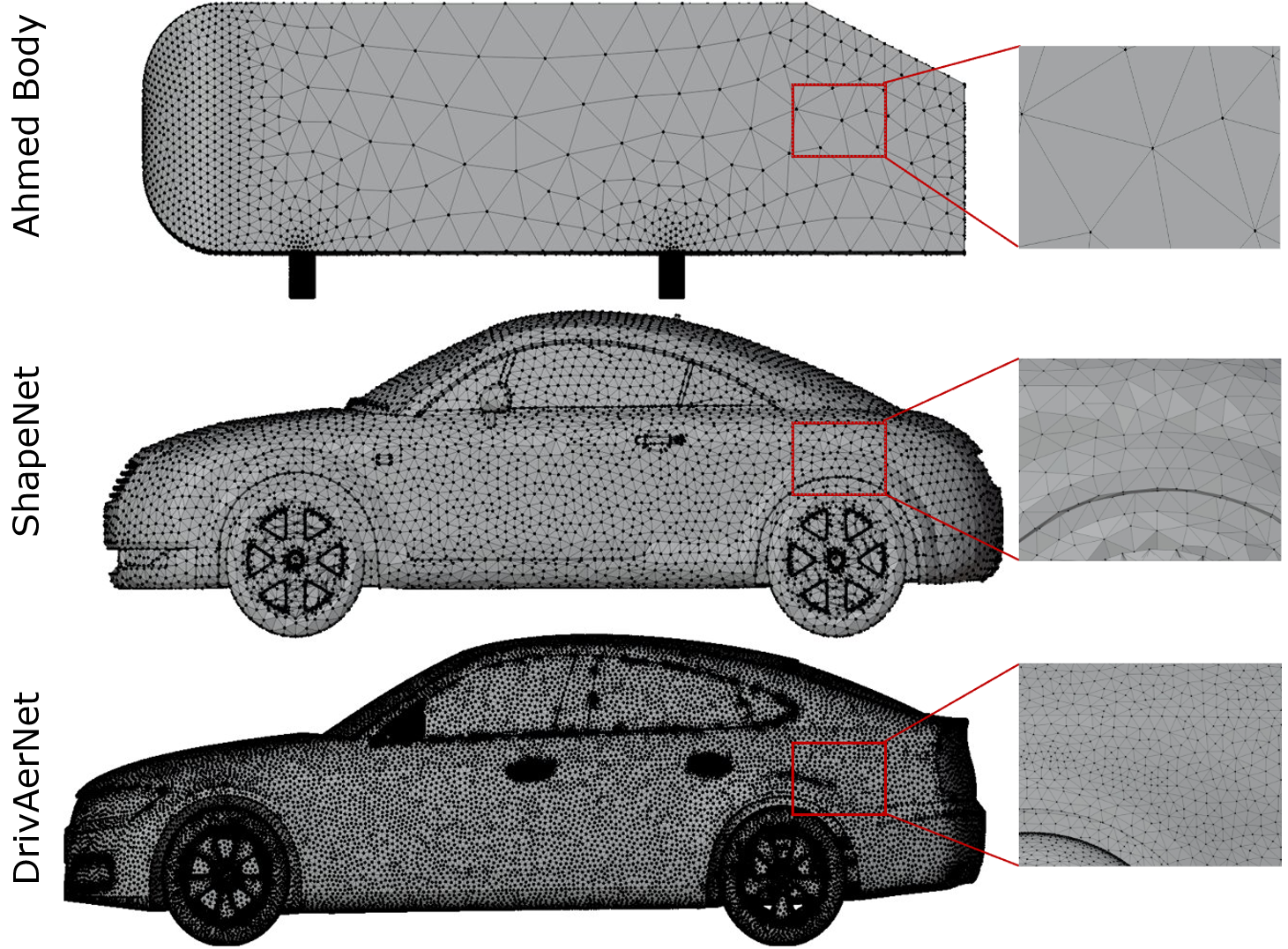}
    \caption{Mesh resolution comparison across various datasets; first row features the Ahmed body mesh from Li et al. (2023)~\cite{li2023geometryinformed}, demonstrating a coarse resolution. Second row shows medium-resolution mesh from the ShapeNet dataset, as utilized by Song et al. (2023)~\cite{song2023surrogate}. Final row presents our high-resolution mesh, providing greater detail for in-depth aerodynamic design.}
    \label{fig:MeshResolutionCompare}
    \vspace{-10pt}
\end{figure}

\begin{figure*}[t!]
        \centering
    \includegraphics[width=0.9\textwidth]{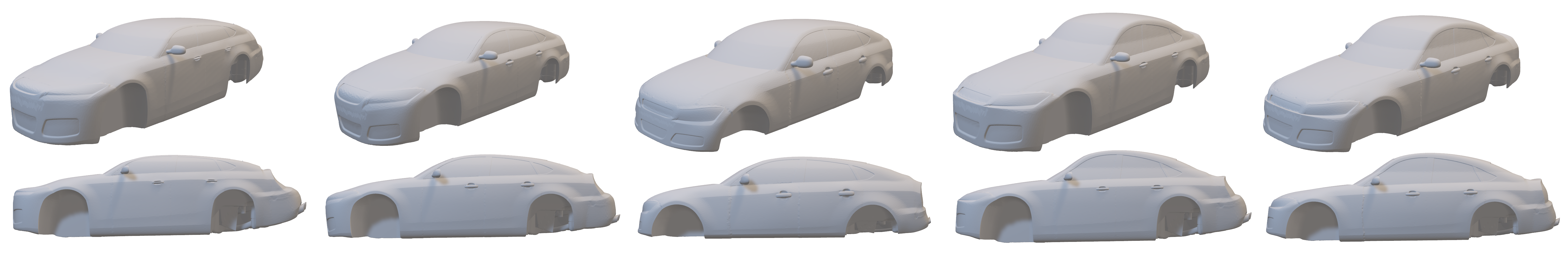}
    \caption{Car models from the DrivAerNet dataset, illustrating a range of aerodynamic designs. The far left model exhibits the maximum volume within the dataset, while the far right model represents the design with the smallest volume, highlighting the diversity and scope of aerodynamic profiles studied.}
    \label{fig:DrivAerPlusPlusSamples}
    \vspace{-10pt} 

\end{figure*}
\vspace{-10pt}
\paragraph{Techniques for Generating Diverse Car Designs:}
The parametric model, along with the constraints and the bounds applied during the Design of Experiments (DoE), significantly enriches the dataset, making it a robust foundation for the development and training of advanced deep learning models aimed at surrogate modeling and design optimization tasks. 
Diverging from the approach taken in~\cite{Jacob2021}, we implemented a broader range of morphing techniques, enabling us to explore a more diverse array of car designs. This approach aims to enhance the adaptability of deep learning models, allowing them to generalize across various car designs instead of being limited to minor geometric modifications within a single design.  Figure~\ref{fig:MeshResolutionCompare} depicts the variation in mesh quality, ranging from coarse to high-resolution across various datasets. Our original mesh with 540k mesh faces, provides a denser and more detailed representation compared to the mesh resolutions in the studies by~\cite{li2023geometryinformed} and~\cite{song2023surrogate}, thereby revealing more detailed geometric and design features.

Additionally, Figure~\ref{fig:DrivAerPlusPlusSamples}
presents a spectrum of car shapes from the DrivAerNet dataset, illustrating the variability in design dimensions and features. This range from the largest to the smallest volume model underscores the dataset's capacity to cover a comprehensive span of aerodynamic profiles.  

\subsection{Numerical Simulation}

\subsubsection{Domain and Boundary Conditions}
The DrivAer fastback model scaled 1:1 was selected for conducting the CFD simulations. The simulations were carried out using the open-source software OpenFOAM®, a comprehensive collection of C++ modules for tailoring solvers and utilities in CFD studies. In this study, the coupling between pressure and velocity was achieved through the SIMPLE algorithm (Semi-Implicit Method for Pressure Linked Equations), as implemented in the simpleFoam solver, which is designed for steady-state, turbulent, and incompressible flow simulations. The $k$-$\omega$-SST model, based on Menter's formulation~\cite{menter2003ten}, was chosen for the Reynolds-Averaged Navier-Stokes (RANS) simulations due to its ability to overcome the limitations of the standard $k$-$\omega$ model, particularly its dependency on the freestream values of $k$ and $\omega$, and its effectiveness in predicting flow separation. 

\begin{figure}[h!]
    \centering
    \includegraphics[width=\columnwidth]{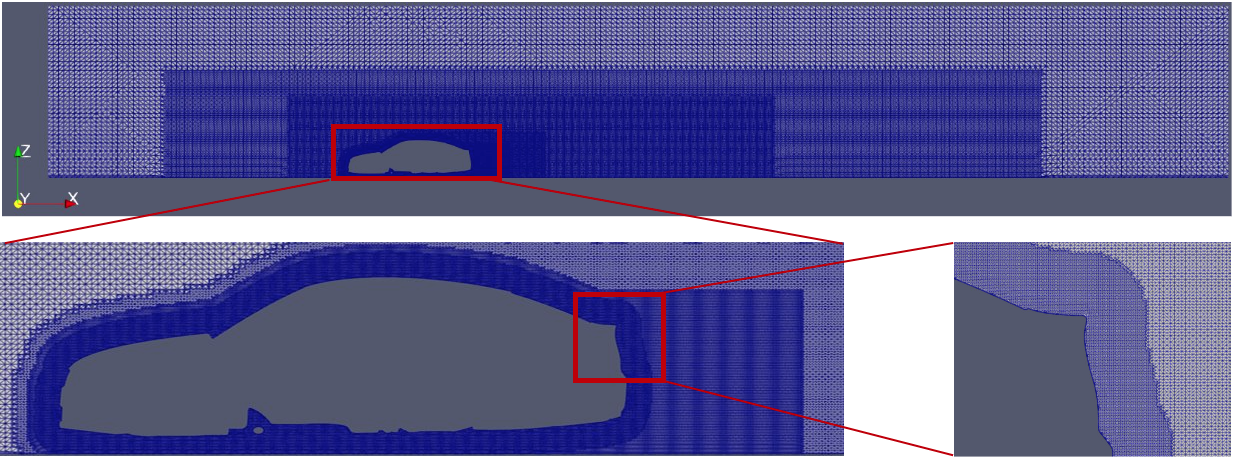}
    \caption{Computational grid in the $y$-normal symmetry plane displays four distinct regions of refinement. The addition of layers around the car body is also evident, all designed to accurately capture both the wake dynamics and the boundary layer development.}
    \label{fig:snappyhexmesh}
    \vspace{-10pt}
\end{figure}

The simulations were performed at a flow velocity ($u_{\infty}$) of 30 m/s, which corresponds to a Reynolds number of roughly $9.39 \times 10^6$, using the car length as the characteristic length scale. The computational mesh was constructed using the SnappyHexMesh (SHM) tool, featuring four distinct refinement zones. Additional layers have been added around the car body to precisely represent wake dynamics and boundary layer evolution (see Figure \ref{fig:snappyhexmesh}).
Boundary conditions were defined with a uniform velocity at the inlet and pressure-based conditions at the outlet. To avoid backflow into the simulation domain, the velocity boundary condition at the outlet was configured as an outflow condition, with a specified inflow condition applied in cases of return flow. The car surface was assigned a no-slip boundary condition, while the wheels were modeled with a rotating wall boundary condition. Slip boundary conditions were applied to the lateral and top boundaries of the domain.

Near-wall viscosity effects were addressed using the \textit{nutUSpaldingWallFunction} wall function approach. The selected wall function for the viscosity term applies a continuous turbulent viscosity profile near the wall, based on velocity, following the approach proposed by~\cite{spalding1974numerical}. 
For divergence terms, the default Gauss linear scheme is used, with the velocity convective term discretized using a bounded Gauss \textit{linearUpwindV} scheme applied to the gradient of velocity, ensuring second-order accuracy.  Gradient calculations employed the Gauss linear method complemented by a multi-dimensional limiter to enhance the stability of the solution. The quantities of interest are the 3D velocity field, surface pressure, and wall-shear stresses, as well as the aerodynamic coefficients. 
\subsubsection{Validation of the Numerical Results}

\begin{table*}[h!]
\centering
\renewcommand{\arraystretch}{1.2} 
\begin{tabular}{cccccc}
    \multicolumn{6}{c}{Comparison of Mesh Resolutions and Experimental/Simulation Results}\\ 
\hline 
Mesh name & Cell count & DrivAer face count & CPU-hours & Difference Exp. & Difference Ref. Sim. \\
\hline 
RANS coarse &$8 \times 10^6$ & $5.4 \times 10^5$ & 88 
& 2.81$\%$& 1.7$\%$\\
RANS fine & $16 \times 10^6$ & $5.4 \times 10^5$ & 205 
&0.81$\%$&1.88$\%$\\
\hline
\end{tabular}
\caption{A comparison of different mesh resolutions used in our simulations for the baseline DrivAer model, including cell count and DrivAer face count. The simulation times are listed alongside the percentage differences when compared to experimental results and a referenced simulation, as reported in~\cite{heft2012experimental}.}
\label{tab:MeshResolution}

\end{table*}
The selection of the DrivAer fastback model is justified by the availability of both computational and experimental references, enabling us to benchmark our results against established data~\cite{heft2012experimental, wieser2014experimental}. Before commencing the simulations, we conducted a preliminary assessment of how mesh refinement influences the results. This involved comparing the drag coefficients obtained from two different mesh resolutions with experimental values and reference simulation, as detailed in Table~\ref{tab:MeshResolution}. The objective was to identify an optimal balance between simulation accuracy and computational efficiency. This balance is crucial because our aim is to generate a large-scale dataset for training deep learning models, necessitating high fidelity in the simulation results, while also ensuring manageable disk storage and simulation times to accommodate the extensive computational requirements.
The drag coefficient, $C_d$, is determined by the equation:
\begin{equation}
C_d = \frac{F_d}{\frac{1}{2} \rho u_{\infty}^2 A_{\text{ref}}}
\end{equation}
The drag force, denoted by $F_d$, experienced by a body is a function of its effective frontal area $A_{ref}$, the freestream velocity $u_{\infty}$, and the air density $\rho$. This force comprises both pressure and frictional components.

The evaluation encompassed not only the drag coefficients but also the mesh size and the computational resources required. 
The simulations were performed on a machine equipped with AMD EPYC 7763 64-Core Processors, totaling 256 CPU cores with 4 Nvidia A100 80GB GPUs.

Our analysis reveals a consistent correlation between our simulations and both the benchmark experimental data and reference simulations. In computational fluid dynamics, especially in the preliminary stages of design, an error margin of up to 5$\%$ is generally considered acceptable for engineering purposes. Therefore, considering the balance between accuracy and computational resource allocation, we decided to conduct our simulations using the 8 million and 16 million cell meshes. These configurations provide a compromise between computational efficiency and the level of detail necessary for accurate aerodynamic analysis.

\paragraph{Rationale for Using RANS Simulations in Large-Scale Dataset Generation:}
While the DrivAerNet dataset provides high fidelity in comparison to existing datasets, it remains limited in scope compared to the industrial standard, where simulations often involve transient hybrid RANS-LES or Wall-Modeled LES models and significantly larger computational grids (e.g., 100-200M cells). It is computationally prohibitive to run 4,000 simulations at this high level of fidelity. Therefore, in practical industrial applications, RANS simulations can be effectively employed to explore the design space during the initial or conceptual design stages. These simulations offer a balance between computational cost and accuracy, skimming the design space efficiently. In later stages of design, higher fidelity simulations, such as hybrid RANS-LES are typically used either for detailed CFD validation or for further training of machine learning models on more complex cases. As shown by \cite{elrefaie2024real}, leveraging multi-fidelity CFD simulations is a robust strategy for accurate 3D flow field estimation. This approach combines the more easily obtainable RANS data, which captures general flow behavior, with high-fidelity Direct Numerical Simulation (DNS) data that provides detailed flow information despite being computationally expensive. Training a deep learning model with such a diverse dataset allows the model to generalize effectively to real-world scenarios, as confirmed by wind tunnel tests. This two-phase training strategy begins by using medium-fidelity RANS data to capture general flow patterns and subsequently fine-tuning the model with DNS data to enhance both accuracy and real-world applicability.

Similar results leveraging multi-fidelity datasets for surrogate model training have been demonstrated by \cite{romor2023multi, shen2024application}, underscoring the efficacy of this approach in aerodynamic analysis. The DrivAerNet dataset can also be utilized in this manner, allowing integration with datasets of either lower or higher fidelity to augment model training and refine predictive capabilities.

\subsubsection{CFD Simulation Results}

\paragraph{Inclusion of Diverse Car Dimensions and Complex Flow Dynamics:}
In contrast to the approach by~\cite{song2023surrogate} where all car models are standardized to a uniform length of 3.5 meters to fit a predefined computational domain, our dataset allows diversity in car dimensions, adjusting the mesh, bounding boxes, and additional layers accordingly for each design. This flexibility is crucial for capturing the intricate flow dynamics around cars, including phenomena like flow separation, reattachment, and recirculation, and ensuring precise aerodynamic coefficient estimation. This approach addresses limitations observed in some studies that prioritize dataset size over simulation fidelity, often overlooking the importance of convergence, accurate modeling, and appropriate boundary conditions for complex 3D models.




\subsection{Geometric Feasibility}

\paragraph{Ensuring Geometric Integrity in Automated Design Morphing:} In our approach to generating large diverse automotive designs through automated morphing with ANSA®, ensuring the geometric quality and feasibility of each variant is crucial. To address potential issues such as non-watertight geometries, surface intersections, or internal holes resulting from the morphing operations, we employ an automated mesh quality assessment and repair process. This procedure not only identifies but also rectifies common geometric anomalies, ensuring that only simulation-ready models are included in our dataset. Geometries failing to meet these criteria are systematically excluded from simulations. The DrivAerNet dataset employs a diverse array of parameters (total of 50 parameters) to morph car geometries, encompassing modifications to the side mirror placement, muffler position, windscreen, rear window length/inclination, the size of the engine undercover, offsets for the doors and fenders, hood positioning, and the scale of headlights as well as alterations to the overall car length and width. Additionally, adjustments are made to the car's upper and underbody scaling, as well as pivotal angles such as the ramp, diffusor, and trunk lid angles, all crucial for exploring the effects of different design modifications on car aerodynamics. For a detailed account of the morphing parameters, including their lower and upper bounds, please refer to our GitHub repository.

As we morph the entire car's geometry, wheel positioning is adjusted in the $x$, $y$, and $z$ axes during the morphing process. For all simulations, we use front and rear wheels of the same shape. To accurately simulate the wheel rotation, we export them as separate STL files post-morphing, which allows us to apply a rotating wall boundary condition. Moreover, morphing affects the car's vertical positioning, necessitating a calculation of the $z$-axis displacement to ensure the car body and wheels are properly aligned with the ground plane. For simulation purposes, we supply three distinct STLs to model their interactions accurately: one for the car body, one for the front wheels, and one for the rear wheels.

\subsection{DrivAerNet Dataset Characteristics}

\begin{figure*}[h!]
    \centering
    \subcaptionbox{Velocity field distribution around the DrivAer model, with velocity magnitudes expressed in (m/s). \label{fig:velocity_DrivAer}}{%
        \includegraphics[width=0.4\textwidth]{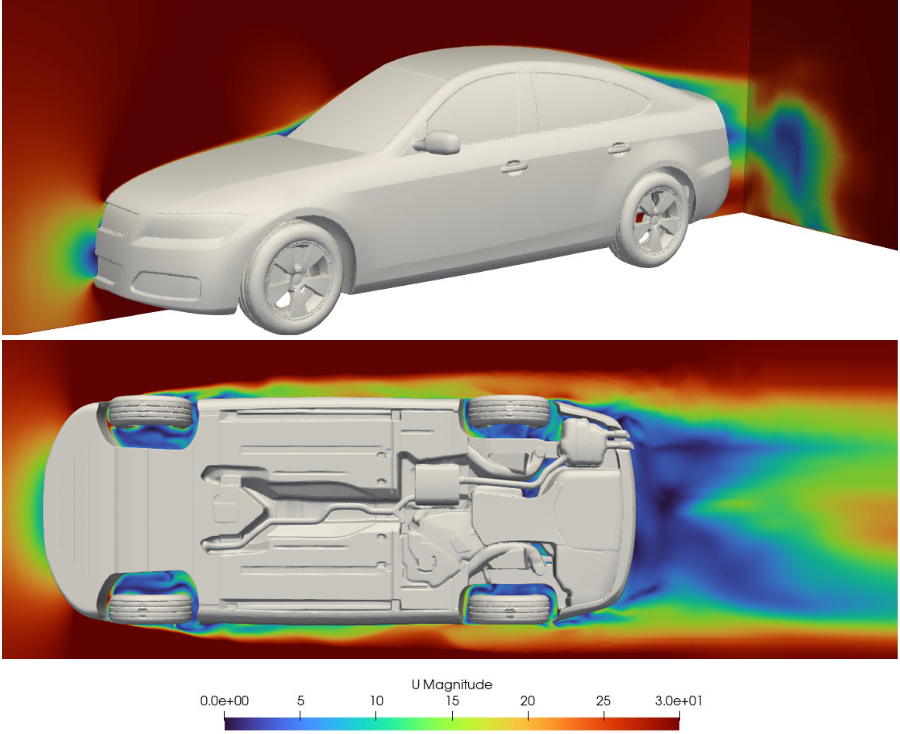}%
    }
\hspace{1em} 
      \subcaptionbox{Pressure coefficient ($C_p$) distribution on the DrivAer model, illustrating the pressure variation across the car's surface.\label{fig:Cp_DrivAer}}{%
        \includegraphics[width=0.4\textwidth]{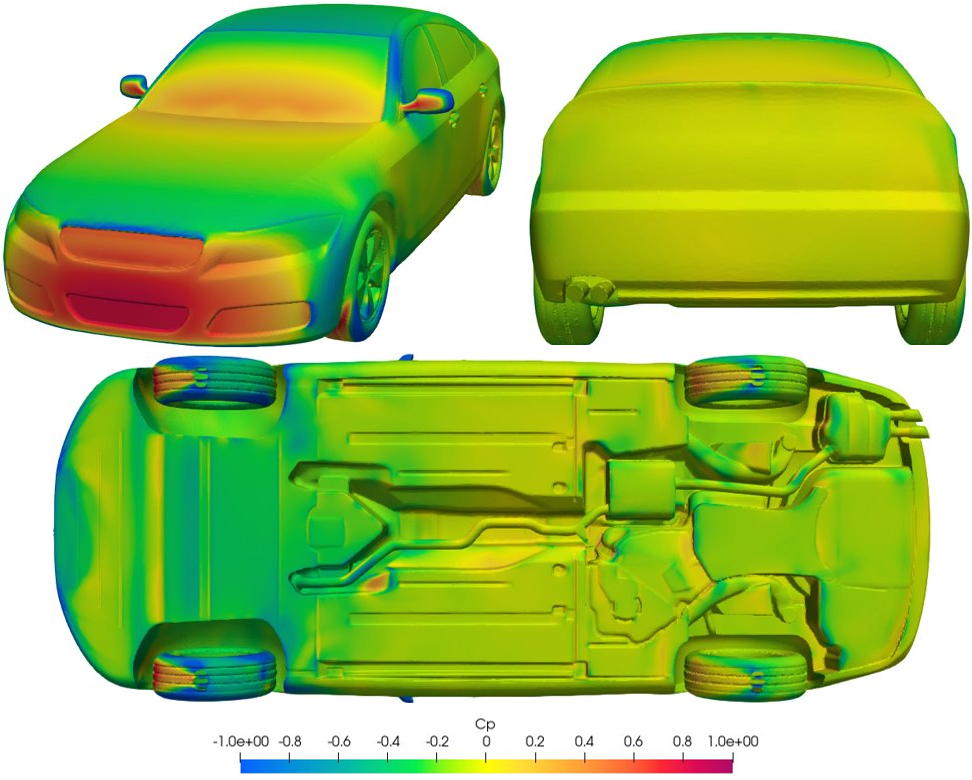}%
    } 
    \caption{The DrivAerNet dataset includes detailed 3D fields for velocity, pressure, and wall-shear stresses, alongside aerodynamic coefficients as well as the detailed 3D meshes of the car bodies and both front and rear wheels in each of its entries.}
    \vspace{-10pt} 

\end{figure*}

For our simulations, we employed OpenFOAM® version 11~\cite{OpenFOAMv11}, executing the computational tasks across 128 CPU cores and 4 Nvidia A100 80GB GPUs. This resulted in a total computational cost of approximately 352,000 CPU hours. We make available the complete suite of data, encompassing both the raw CFD outputs and the derived post-processed datasets.

Our dataset serves as a benchmark for evaluating deep learning models, designed to facilitate effective model testing. To manage the large volumes of data from CFD simulations, we addressed dataset size and accessibility by dividing DrivAerNet into smaller subsets tailored to specific tasks, reducing storage needs. For example, the 4000 car geometries in STL format occupy 84 GB, while aerodynamic performance coefficients, including the total moment coefficient \(C_m\), total drag coefficient \(C_d\), total lift coefficient \(C_l\), front lift coefficient \(C_{l,f}\), and rear lift coefficient \(C_{l,r}\), are stored in CSV format. The full CFD volumetric field, which contains velocity, pressure, and wall-shear stress data, occupies 5.6 TB in VTK format. Surface pressure field data in VTK format totals 102 GB, and wall shear stress (WSS) data sums up to 200 GB, also in VTK format. We have also partnered with Harvard Dataverse to provide easy access to these subsets, allowing researchers to download only the data they need, thereby minimizing storage requirements. A GitHub repository will be updated with detailed subset information and download guidelines. Given the widespread use of data visualization tools such as ParaView and VisIt, which rely on the Visualization Toolkit (VTK), our dataset is made available in VTK format. This ensures that the data is easily accessible and usable within these common visualization environments, supporting a broad range of research and application needs.

The DrivAerNet dataset features full 3D flow field information, as shown in Figure~\ref{fig:velocity_DrivAer} with the velocity data, and in addition, it provides the pressure distribution on the car's surface. The pressure coefficient, denoted as \( C_p \), is calculated as the ratio of the pressure differential \( p - p_{\infty} \) to the dynamic pressure, \( \frac{1}{2} \rho  u_{\infty}^2 \), and is given by:
\begin{equation}
C_p = \frac{p - p_{\infty}}{\frac{1}{2} \rho u_{\infty}^2}.
\end{equation}
The distribution of \( C_p \) on the car surface is depicted in Figure~\ref{fig:Cp_DrivAer}.

The DrivAerNet dataset offers a comprehensive suite of aerodynamic data pertinent to car geometries, including key metrics such as the total moment coefficient \(C_m\), the total drag coefficient \(C_d\), total lift coefficient \(C_l\), front lift coefficient \(C_{l,f}\), and rear lift coefficient \(C_{l,r}\). Included within the dataset are crucial parameters like wall-shear stress and the $y^{+}$ metric, integral for mesh quality evaluations. Further, the dataset provides insights into flow trajectories and detailed cross-sectional analyses of pressure and velocity fields along the $x$ and $y$-axes, enriching the understanding of aerodynamic interactions. 

An overview of the dataset components can be found in Table~\ref{tab:DrivAerNet_Characteristics}, which details the types, formats, and information provided for each dataset element, including the 3D car geometries, aerodynamic performance coefficients, full CFD volumetric fields, and surface fields like pressure and wall-shear stresses.

\begin{table*}[h!]
\renewcommand{\arraystretch}{1.5}
\centering
\begin{tabular}{p{6cm}p{2cm}p{7cm}}
\hline
\textbf{Dataset Component} & \textbf{Format/Type} & \textbf{Information Provided} \\ \hline
3D Car Geometries & STL & Surface representation of the car geometries \\ \hline
Aerodynamic Performance Coefficients  & CSV & Total moment coefficient \(C_m\),  total drag coefficient \(C_d\), total lift coefficient \(C_l\), front lift coefficient \(C_{l,f}\), and rear lift coefficient \(C_{l,r}\) \\ \hline
Full CFD Volumetric Field & VTK & Velocity field, pressure field, wall-shear stresses \\ \hline
Surface Pressure Field & VTK  & Pressure distribution on the car surface \\ \hline
Wall Shear Stresses & VTK & Wall-shear stresses on the car surface  \\ \hline
\end{tabular}
\caption{Overview of the DrivAerNet dataset components, including 3D car geometries, full CFD volumetric fields, surface pressure fields, and wall shear stresses. The dataset provides detailed information on aerodynamic performance, facilitating the analysis of car designs and their aerodynamic characteristics, especially focusing on how design variations impact the flow field and aerodynamic efficiency.}
\label{tab:DrivAerNet_Characteristics}
\end{table*}

\begin{figure}[h!]
    \centering
\includegraphics[width=0.9\columnwidth]{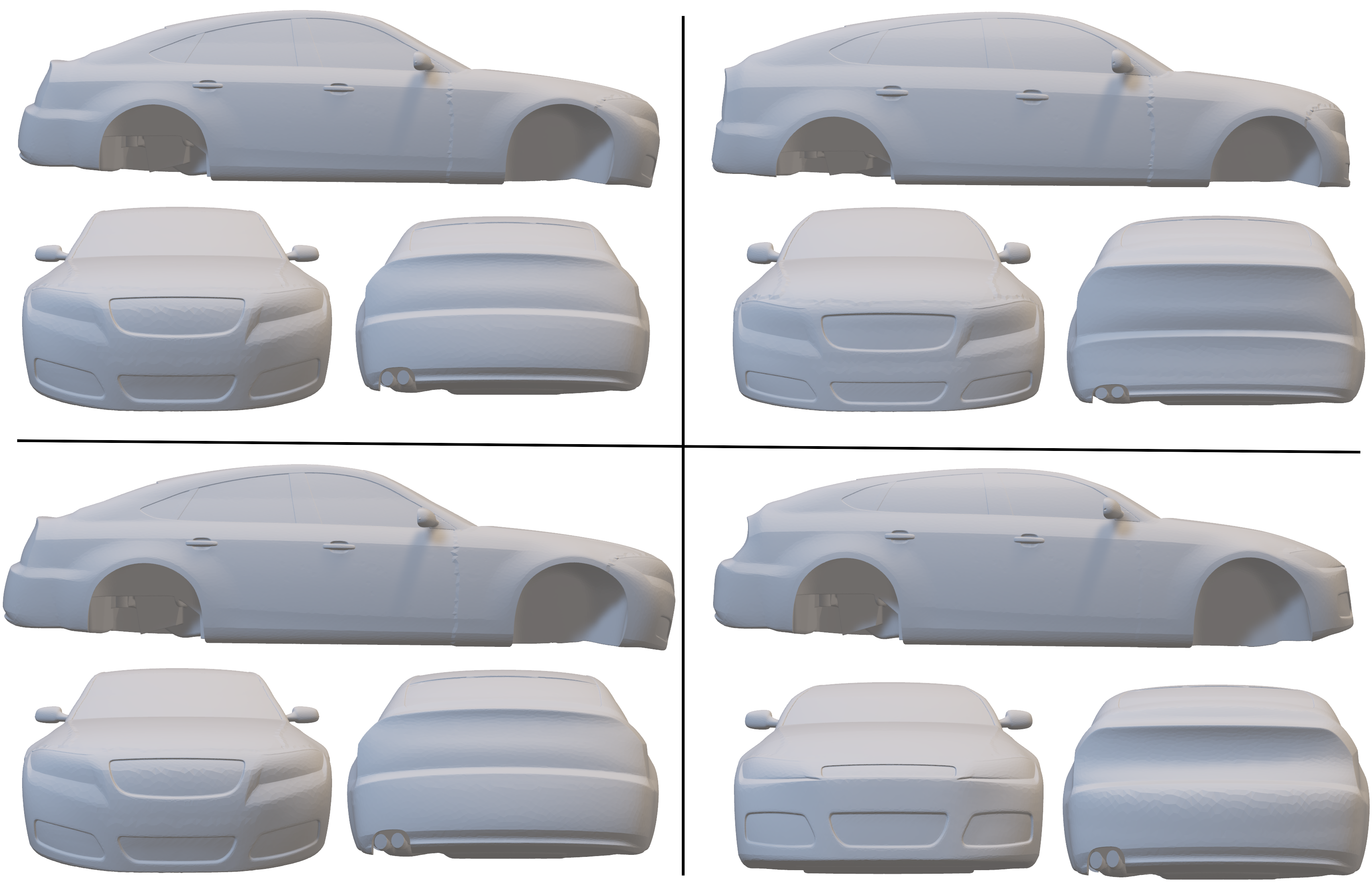}
    \caption{Aerodynamic performance of car designs from DrivAerNet showcasing a range of coefficients. Top left: Design with minimum drag coefficient $C_d$, indicating optimal aerodynamic efficiency. Top right: Design with maximum $C_d$. Bottom left: Design with minimum lift coefficient $C_l$ (highest downforce). Bottom right: Design with maximum $C_l$.}
    \label{fig:DrivAerDesignsDragLift}
    \vspace{-20pt}
\end{figure}

\paragraph{Aerodynamic Performance Variability Amongst Car Designs in DrivAerNet:}
In Figure ~\ref{fig:DrivAerDesignsDragLift}, we present the aerodynamic performance across various designs. The top left illustrates the design with the lowest drag coefficient $C_d$. In contrast, the top right shows the design with the highest $C_d$, identifying opportunities for aerodynamic refinement. The bottom left reveals the design with the lowest lift coefficient $C_l$ (indicating the largest downforce), which is advantageous for stability at high speeds, while the bottom right exposes the design with the highest $C_l$, potentially complicating aerodynamic stability. 

\begin{figure*}[h!]
    \centering
    \includegraphics[width=0.8\textwidth]{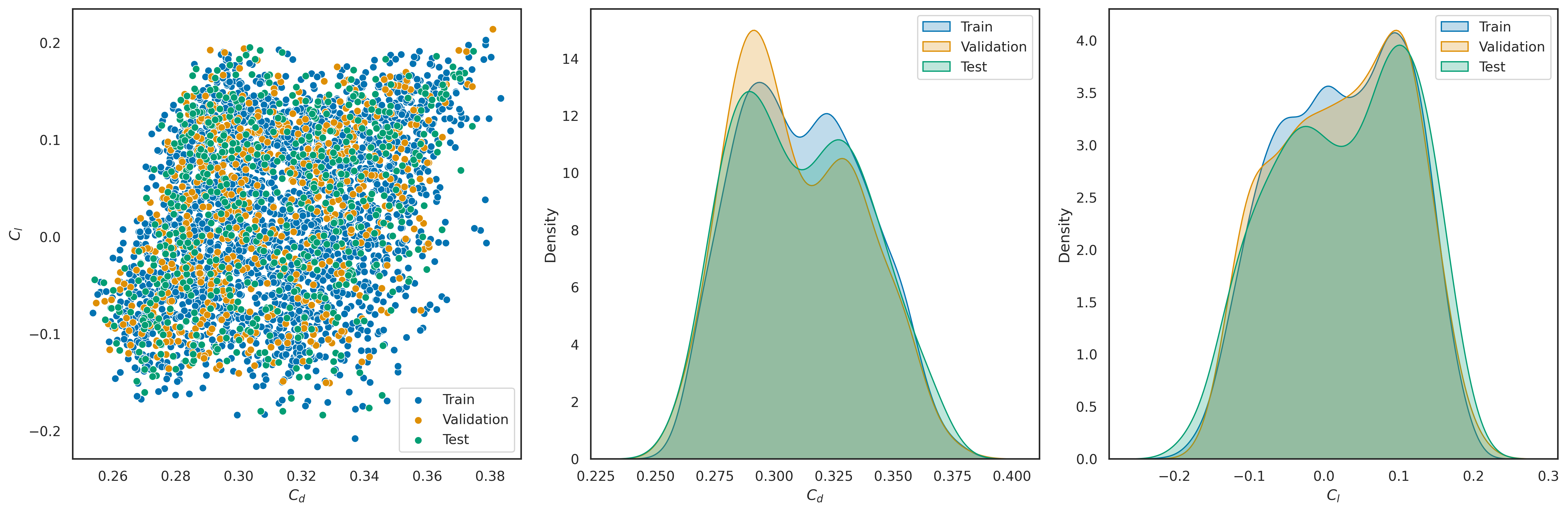}
    \caption{Scatter and kernel density estimation plots illustrating the relationship between drag coefficient ($C_d$) and lift coefficients ($C_l$) for the DrivAerNet dataset. Data points, representing unique design variants generated via Optimal Latin Hypercube sampling and the Enhanced Stochastic Evolutionary Algorithm (ESE), are categorized into training, validation, and test sets (70$\%$, 15$\%$, 15$\%$).}
    \label{fig:pairplotAeroCoefficients_DrivAer}
\end{figure*}

Figure~\ref{fig:pairplotAeroCoefficients_DrivAer} illustrates scatter and kernel density estimation plots showing the relationships between the drag coefficient (\(C_d\)) and lift coefficients (\(C_l\)) for the DrivAerNet dataset. The data points represent unique design variants generated via Optimal Latin Hypercube sampling and the Enhanced Stochastic Evolutionary Algorithm (ESE). The dataset is categorized into training, validation, and test sets, with a division of 70$\%$ for training and 15$\%$ each for validation and testing. This division is critical for maintaining the integrity of the model training process and ensuring robust performance evaluation.

\paragraph{Analysis of Datasets Diversity Using Chamfer Distance}

The Chamfer distance is a metric used to quantify the geometric diversity between two point sets, making it highly applicable for analyzing the diversity in datasets of 3D models~\cite{fan2017point}. It is defined as follows:
\begin{equation}
CD(A, B) = \frac{1}{|A|} \sum_{a \in A} \min_{b \in B} \|a - b\|^2 + \frac{1}{|B|} \sum_{b \in B} \min_{a \in A} \|a - b\|^2
\end{equation}
Where:
\begin{itemize}
    \item \( \|a - b\|^2 \) represents the squared Euclidean distance between point \(a\) from set \(A\) and point \(b\) from set \(B\).
    \item \( \min_{b \in B} \|a - b\|^2 \) is the minimum squared distance from any point \(a\) in \(A\) to points in \(B\).
    \item \( \min_{a \in A} \|b - a\|^2 \) is the minimum squared distance from any point \(b\) in \(B\) to points in \(A\).
\end{itemize}

This measure allows for the calculation of a diversity score by averaging pairwise Chamfer distances across all model pairs within a dataset, offering a quantitative measure of model dissimilarity. We computed the diversity scores for both the Song et al.~\cite{song2023surrogate} dataset (based on ShapeNet) and DrivAerNet datasets. ShapeNet achieved a diversity score (\(CD\)) of 0.433, demonstrating a wide range of car shapes including trucks, sports cars, and various utility cars, which introduces substantial shape variation. Conversely, DrivAerNet recorded a diversity score of 0.171, concentrating on more precise geometries tailored to conventional car designs. This targeted focus aids in enhancing model performance by enabling detailed learning from subtle geometric variations within a more restricted range of automotive shapes.

\begin{figure}[h!]
    \centering
    \includegraphics[width=0.9\columnwidth]{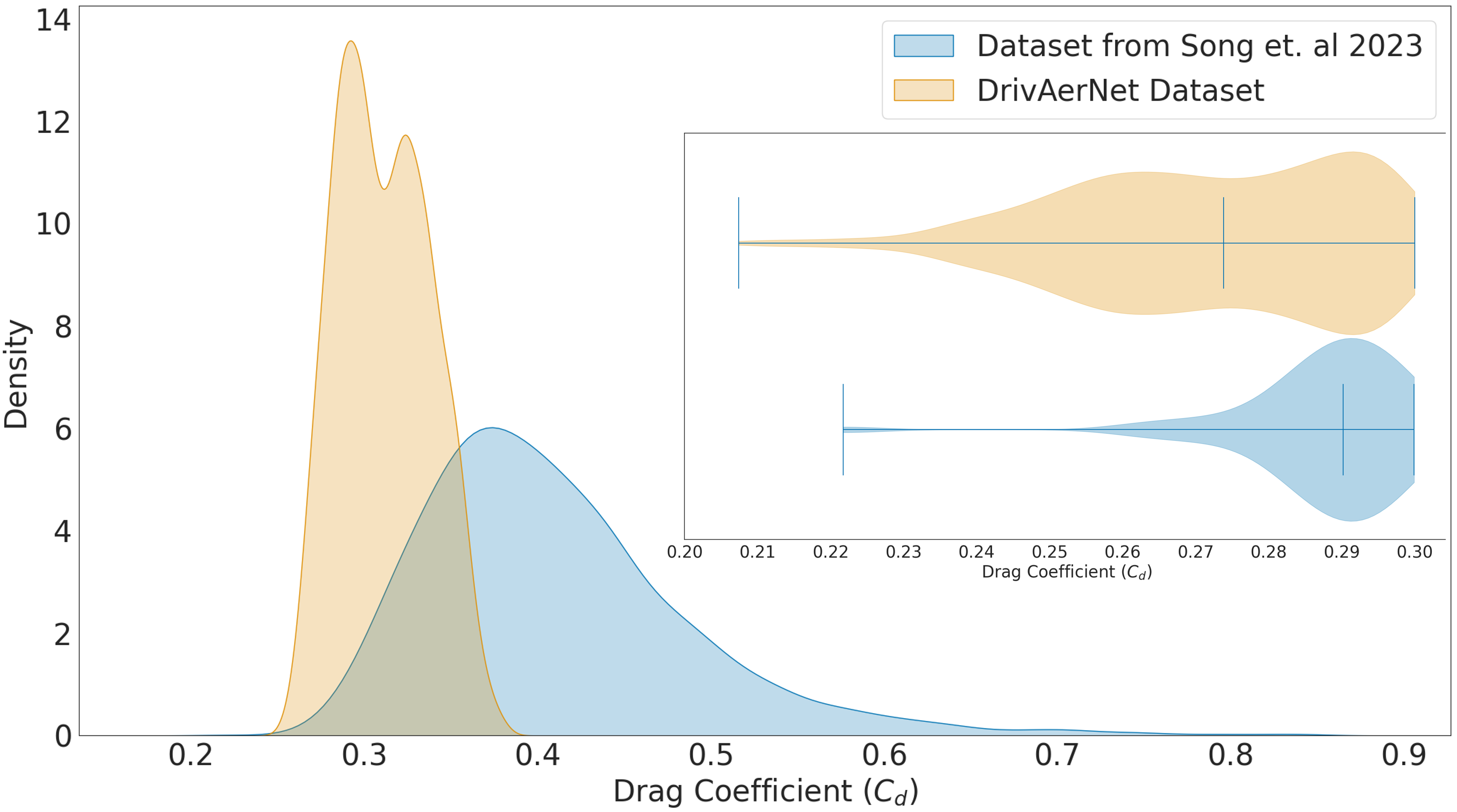}
    \caption{Comparative Kernel Density Estimation (KDE) and violin plots of drag coefficients from two aerodynamic datasets. The blue curve represents the dataset from Song et al. 2023~\cite{song2023surrogate}, and the orange curve corresponds to the DrivAerNet dataset. DrivAerNet focuses on conventional car designs, emphasizing the influence of minor geometric modifications on aerodynamic efficiency.}
    \label{fig:KDE_Datasets}
    \vspace{-10pt}
\end{figure}

This concept of diversity is crucial when examining the practical impacts of these variations on aerodynamic efficiency, as depicted in the Kernel Density Estimation (KDE) plot shown in Figure~\ref{fig:KDE_Datasets}. The KDE plot compares the distribution of drag coefficients between the broad-ranging ShapeNet dataset and our more specialized DrivAerNet dataset. While ShapeNet covers a broad spectrum of drag values reflecting its diverse car designs, DrivAerNet focuses on conventional car designs with detailed geometric modifications, pertinent to the engineering design process. In this process, an initial car design is incrementally refined to optimize aerodynamic performance. Therefore, the DrivAerNet dataset provides an in-depth look at how subtle design changes affect aerodynamic performance, underscoring the practical application of diversity scoring in real-world engineering scenarios. While ShapeNet offers design diversity, its poor mesh quality and geometry issues make it unsuitable for high-fidelity CFD simulations. In contrast, DrivAerNet's parametric design ensures high-quality, automated meshing, leading to more accurate and reliable CFD results, even with a more focused range of designs.

In the next section, we propose a new deep learning method to utilize the dataset for surrogate modeling.
\section{Dynamic Graph Convolutional Neural Network for Regression}
\label{sec:RegDGCNN}
\begin{figure*}[ht!]
    \centering
    \includegraphics[width=0.8\textwidth]{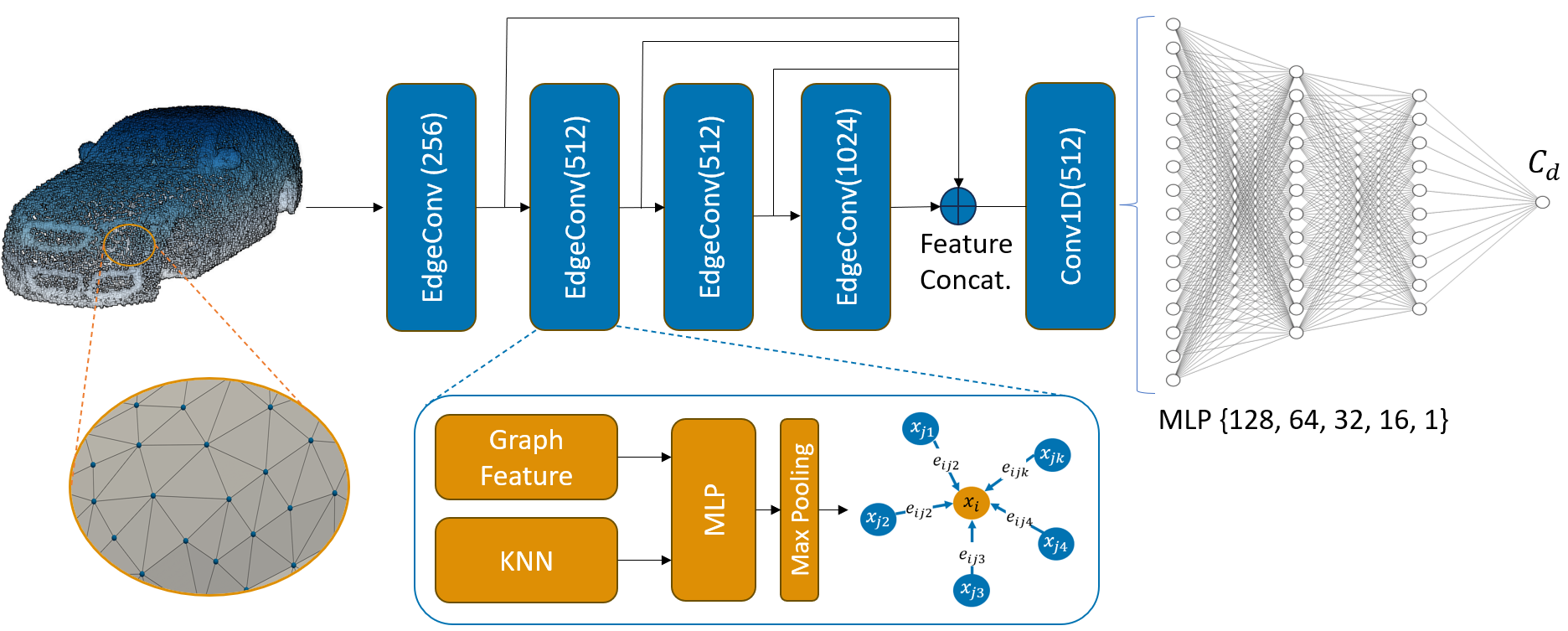}
\caption{Architecture of RegDGCNN for aerodynamic drag prediction. The model processes a 3D mesh by converting it into a point cloud representation. It takes $n$ input points, calculates an edge feature set of size $k$ for each point at an EdgeConv layer, and aggregates features within each set to compute EdgeConv responses for the corresponding points. The output features of the last EdgeConv layer are aggregated globally to form a 1D global descriptor, which is then used to predict the aerodynamic drag coefficient $C_d$, enabling direct learning from the 3D geometry of the object. The EdgeConv block ingests an input tensor of dimensions \( n \times f \), where it determines edge features for each point utilizing a multi-layer perceptron (MLP). Post-MLP application, the block outputs a tensor of dimensions \( n \times a_n \) by conducting a pooling operation over the neighboring edge features.
}
    \label{fig:RegDGCNN}
    \vspace{-10pt}
\end{figure*}

Geometrical deep learning has demonstrated significant promise in addressing fluid dynamics challenges involving irregular geometries, as demonstrated in studies by 
\cite{sanchez2020learning, pfaff2020learning,
abbas2022geometrical, kashefi2022physics, Rios2019, Rios2021_Point2FFD,Rios2019_EfficiencyPointClouds}.

In this work, we extend the Dynamic Graph Convolutional Neural Network (DGCNN) framework~\cite{wang2019dynamic}, traditionally associated with PointNet~\cite{qi2017pointnet} and graph CNN methodologies, to address regression tasks, marking a significant departure from its conventional applications in classification. Our contribution lies in adapting the DGCNN architecture to predict continuous values, specifically focusing on aerodynamic coefficients critical in fluid dynamics and engineering design. Leveraging the spatial encoding capabilities of PointNet and the relational inferences provided by graph CNNs, our proposed RegDGCNN model (as shown in Figure~\ref{fig:RegDGCNN}) aims to capture the complex interactions of fluid flow around objects, offering a novel method for accurate estimation of crucial aerodynamic parameters.
This approach harnesses local geometric structures by constructing a local neighborhood graph and applying convolution-like operations on the edges linking pairs of neighboring points, aligning with graph neural network principles. The technique, termed edge convolution (EdgeConv), is shown to exhibit properties that bridge translation invariance and non-locality. Uniquely, unlike in standard graph CNNs, the graph of RegDGCNN is not static but is dynamically updated after each layer of the network, allowing the graph structure to adapt to evolving feature spaces. This dynamic graph construction results in better performance for 3D geometric data while maintaining computational efficiency during both training and inference, as the graph is built on top of the sampled point clouds. Moreover, DGCNN is well-suited for handling unstructured point cloud data, which aligns well with the nature of the mesh format in our dataset.

We start by initializing a graph $G$ with node features $X$, along with the parameters for the EdgeConv layers $\theta$ and the Fully Connected (FC) layers $\phi$. A distinctive feature of the RegDGCNN is its dynamic graph construction within each EdgeConv layer, where the $k$-nearest neighbors of each node are identified based on the Euclidean distance in the feature space, thereby adaptively updating the graph's connectivity to reflect the most significant local structures. The EdgeConv operation, defined as:
\begin{equation}
    h_{ij} = \Theta \left( x_i, x_j - x_i \right)
\end{equation}
enhances node features by aggregating information from these neighbors using a shared Multi-Layer Perceptron (MLP), which processes both the individual node features and their differences with adjacent nodes, capturing the local geometric context effectively. For the EdgeConv layers, we use same convolution with a kernel size of 1, meaning the input and output dimensions remain unchanged throughout each EdgeConv operation. This helps maintain the feature space dimensions while effectively capturing local geometric features through the dynamic graph structure.

Following the EdgeConv transformations, global feature aggregation is performed, pooling features from across all nodes into a singular global feature vector:
\begin{equation}
    x'_i = \max_{j \in \mathcal{N}(i)} h_{ij}
\end{equation}
Here, max pooling is employed to encapsulate the graph's holistic information. This global feature vector is subsequently processed through several FC layers, with the inclusion of non-linear activation functions like LeakyReLU and dropout to introduce non-linearity and prevent overfitting, respectively. The architecture culminates in an output layer, designed to suit the specific task at hand, for instance, employing a linear activation for regression tasks.
\begin{equation}
    h_{ij} = \text{MLP}\left( \left[ x_i, x_j - x_i \right] \right), \quad
        X' = \max_{i \in \mathcal{G}} x'_i
\end{equation}

The model's performance is quantified by calculating the loss between its predicted outputs and the ground truth drag values using the mean squared error (MSE), with the backpropagation algorithm adjusting the model parameters $\theta$ and $\phi$ through optimization algorithms such as Adam~\cite{kingma2014adam} to minimize this loss. This iterative refinement process highlights RegDGCNN's capability to dynamically leverage and integrate hierarchical features from graph-structured data.

\subsection{Implementation Details}
\paragraph{Network Architecture: } We constructed the graph for RegDGCNN using the $k$-nearest neighbors algorithm, with $k$ set to 40. This parameter was critical in defining the local neighborhood over which convolutional operations were performed.  The RegDGCNN model was instantiated with specific parameters to accommodate the nature of the regression task. The EdgeConv layers were configured with channels of sizes \{256, 512, 512, 1024\}, and the following MLP layers were \{128, 64, 32, 16\}. Lastly, the embedding dimension of the network was set at 512, providing a high-dimensional space to capture the complex features necessary for the regression tasks at hand. Our RegDGCNN model is entirely differentiable and can seamlessly integrate with 3D generative AI applications for enhancing design optimization.
\vspace{-10pt}
\paragraph{Model Hyperparameters: } The experiments were conducted using the PyTorch framework~\cite{paszke2017automatic}. The model was trained with a batch size of 32 and each input containing 5,000 points. The training was distributed across four NVIDIA A100 80GB GPUs using data parallelism for improved computational efficiency, and the process took approximately 3.2 hours to complete. The network's learning rate was initially set to 0.001, and a learning rate scheduler was employed to reduce the rate upon plateauing of the validation loss, specifically using the \textit{ReduceLROnPlateau} scheduler with a patience of 10 epochs and a reduction factor of 0.1. This approach helped in fine-tuning the model by adjusting the learning rate in response to the performance on the validation set. The model was trained for a total of 100 epochs, ensuring sufficient learning while preventing overfitting. For optimization, we used the Adam optimizer~\cite{kingma2014adam} due to its adaptive learning rate capabilities.

\paragraph{Evaluation Metrics}

Various loss functions are used to evaluate the accuracy of models in predicting the aerodynamic drag of different cars. The primary metrics are described below:
\begin{itemize}
    \item \textbf{Mean Squared Error (MSE):}
This metric calculates the average of the squared discrepancies, which represent the differences between the actual drag coefficients obtained from CFD simulations and those predicted by the surrogate models. It is particularly responsive to large deviations in predictions.
\begin{equation}
    MSE = \frac{1}{n} \sum_{i=1}^{n} (C_{d_{i}} - \hat{C}_{d_{i}})^2
\end{equation}
\item \textbf{Coefficient of Determination (\(R^2\) Score):} The \(R^2\) Score quantifies the fraction of variance in the observed drag coefficients that can be accounted for by the model's predictions. An \(R^2\) value of 1 indicates a perfect prediction.
\begin{equation}
    R^2 = 1 - \frac{\sum_{i=1}^{n} (C_{d_{i}} - \hat{C}_{d_{i}})^2}{\sum_{i=1}^{n} (C_{d_{i}} - \bar{C}_{d})^2}
\end{equation}

\end{itemize}

\section{Surrogate Modeling of Aerodynamic Drag}
\label{sec:SurrogateModeling}

In this section, we evaluate our RegDGCNN model on two aerodynamic datasets, DrivAerNet and ShapeNet, and highlight the impact of larger training volumes on model performance.

\begin{figure}[b!]
    \centering
    \includegraphics[width=0.9\columnwidth]{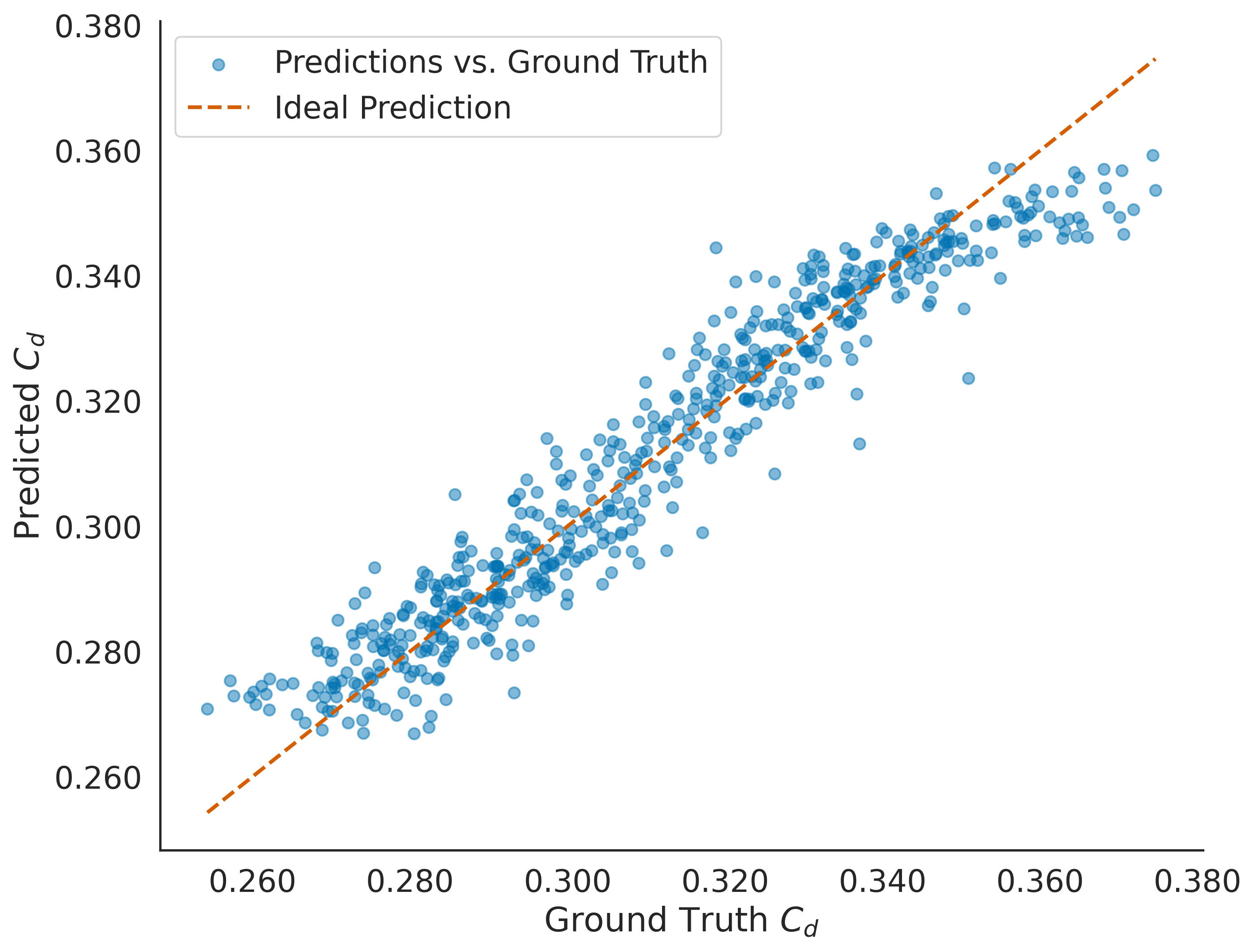}
\caption{Correlation plot of the predicted drag coefficient \(C_d\) by our RegDGCNN model against the ground truth for the DrivAerNet unseen test set, achieving an \(R^2\) score of 0.9. The dotted line represents perfect correlation, indicating the ideal prediction scenario.}
\label{fig:CorrelationPlotDrivAerNet}
\end{figure}
\subsection{DrivAerNet: Aerodynamic Drag Prediction of High-Resolution Meshes}
The examination of RegDGCNN's performance on the DrivAerNet dataset, depicted in Figure~\ref{fig:CorrelationPlotDrivAerNet}, reveals a good correlation between predicted values and ground-truth data from CFD, underscoring the model's effectiveness. The complexity of the DrivAerNet dataset is attributed to its inclusion of industry-standard shapes, varied through 50 geometric parameters, presenting a comprehensive challenge in aerodynamic prediction.
Our model effectively navigated the complexities of the dataset and directly processed the 3D mesh data, marking a significant shift from traditional methods that often rely on generating Signed Distance Fields (SDF) or rendering 2D images. This straightforward approach enabled us to achieve an \(R^2\) score of 0.9 on the unseen test set, emphasizing the model's ability to accurately discern subtle aerodynamic differences.

\subsection{ShapeNet: Aerodynamic Drag Prediction of Arbitrary Car Shapes}

\begin{figure}[h!]
    \centering
    \includegraphics[width=0.9\columnwidth]{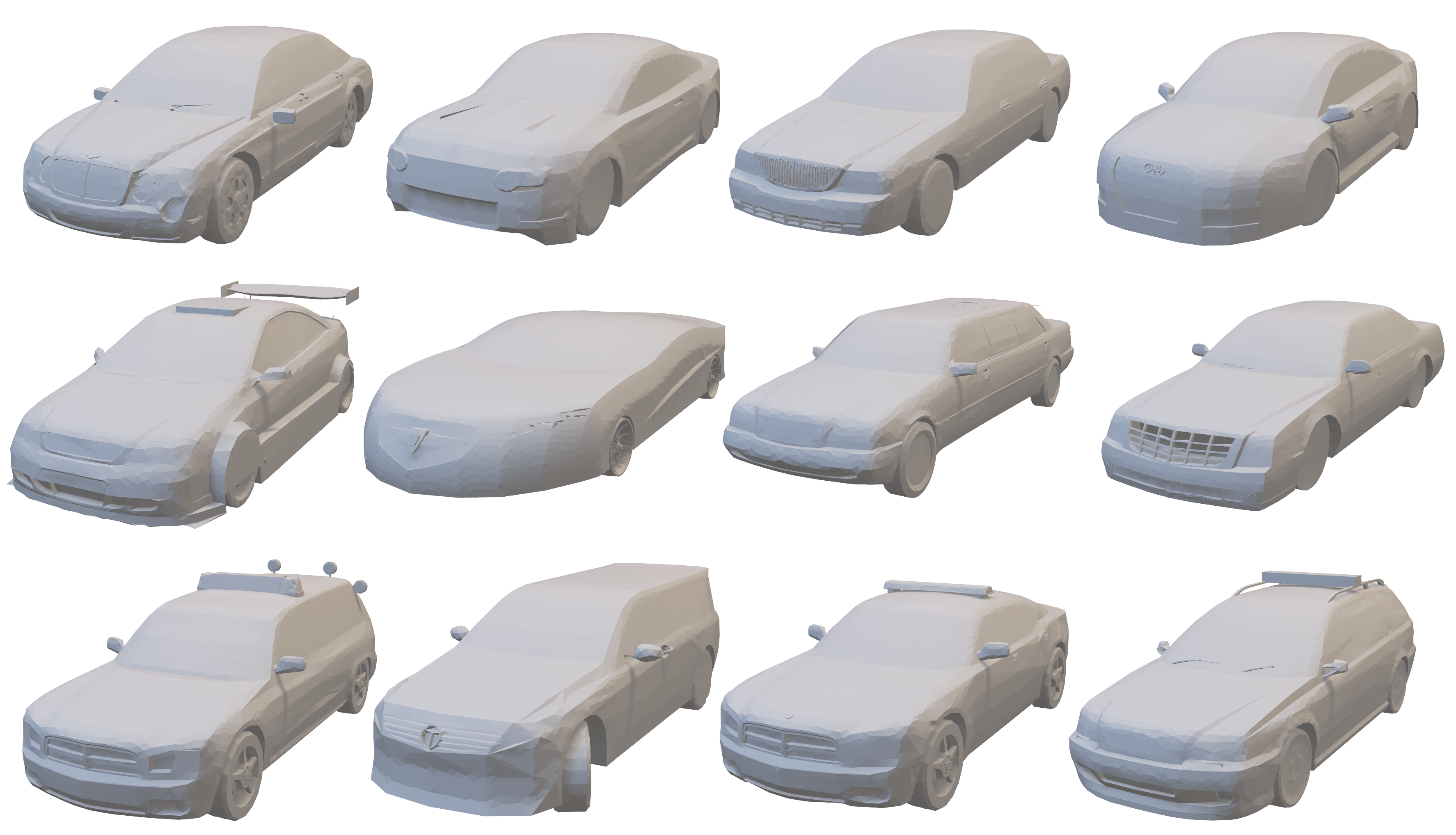}

\caption{A selection of car samples from the ShapeNet dataset, demonstrating the diversity in car shapes and mesh resolutions, utilized to evaluate the generalization capabilities of RegDGCNN. These samples provide a comparative baseline against the high-resolution meshes found in our DrivAerNet dataset.}
    \label{fig:ShapeNetSamples}
    \vspace{-8pt}
\end{figure}

To test the generalizability of the proposed RegDGCNN model, we also evaluated its ability to adapt to complex geometries on an existing benchmark dataset, utilizing 2,479 diverse car designs from the ShapeNet dataset \cite{song2023surrogate} (see Figure~\ref{fig:ShapeNetSamples}), which exhibits a broader range of car shapes than our DrivAerNet dataset. 

In Table~\ref{tab:R2ShapeNet}, we compare the performance of two models: the attn-ResNeXt model from the study by \cite{song2023surrogate}, which implements a self-attention mechanism to boost the understanding of interactions among various regions of an image. It uses 2D depth/normal renderings as input and has approximately 2 billion parameters, achieving an \( R^2 \) score of 0.84; our proposed RegDGCNN model, which directly processes 3D mesh data, significantly reduces the number of parameters to 3 million, and achieves a superior \( R^2 \) score of 0.87. The superior performance of RegDGCNN compared to attn-ResNeXt can be attributed to the fact that the model uses 1000x fewer parameters and works directly on the 3D geometry rather than 2D rendered depth/normal maps, which allows it to learn more features and helps to avoid overfitting. This comparison underscores the efficiency and effectiveness of our model in aerodynamic drag prediction tasks.

\begin{table}[ht]
\centering
\caption{Comparison of model performance for drag prediction on arbitrary car shapes from the ShapeNet dataset.}
\label{tab:model_comparison}
\footnotesize
\setlength{\tabcolsep}{2.5pt} 
\renewcommand{\arraystretch}{1.3} 
\begin{tabular}{lccc}
\hline
Model & Input & \( R^2 \) & $\#$ Parameters \\
\hline
attn-ResNeXt~\cite{song2023surrogate} & 2D Depth/Normal Renderings
 & 0.84 & 2B \\
RegDGCNN (ours)  & 3D Mesh & \textbf{0.87} & \textbf{3M} \\
\hline
\end{tabular}
\label{tab:R2ShapeNet}
\end{table}

\subsection{Effect of Training Dataset Size}
\label{sec:TrainingSize}
For both ShapeNet\footnote{The work expanded their dataset to 9,896 variations through resizing and flipping augmentations but considered only 2,474 unique car designs as independent samples for their surrogate model training to prevent data leakage across different sets.} and DrivAerNet datasets, we first allocated 70$\%$ for training, and 15$\%$ each for validation and testing. Subsequently, we experimented with training subsets at 20$\%$, 40$\%$, 60$\%$, 80$\%$, and 100$\%$ of the training portion. The ShapeNet subsets ranged from 1270 to 6352 samples. Meanwhile, for the DrivAerNet dataset, the corresponding sample sizes were  560, 1120, 1680, 2240, and 2800 samples.

\begin{figure}[t!]
    \centering
    \includegraphics[width=\columnwidth]{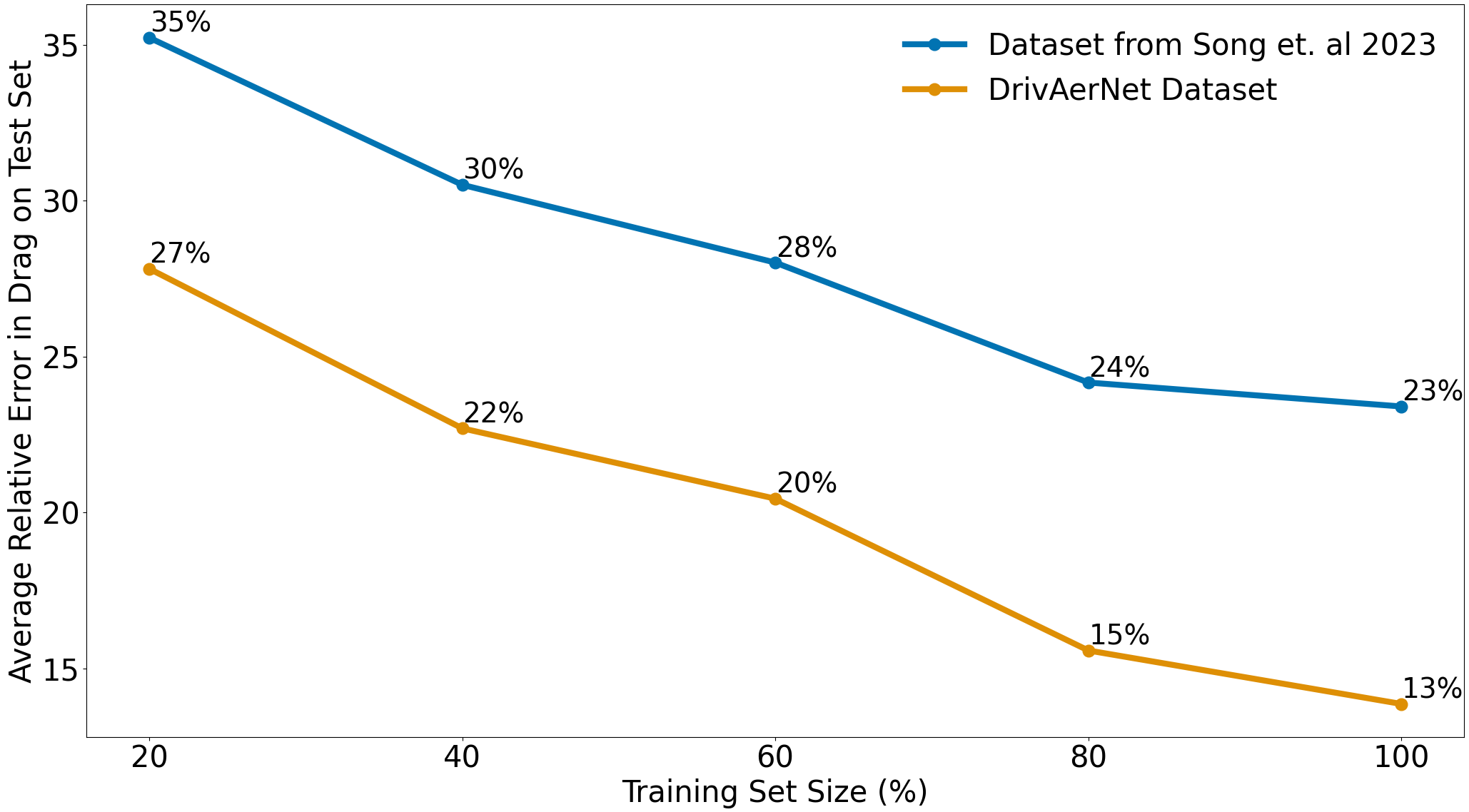}
    \caption{Average relative error in drag coefficient predictions on the unseen test set for our model RegDGCNN, based on training set size. The results from the ShapeNet drag dataset~\cite{song2023surrogate} are denoted in blue, while results from the DrivAerNet dataset are shown in orange. The training set sizes vary from 20$\%$ to 100$\%$. We observe that the increase in dataset size leads to significant error reduction, showing the necessity of larger datasets in aerodynamic surrogate modeling.}
    \label{fig:DatasetSize}
    \vspace{-10pt}
\end{figure}

Figure~\ref{fig:DatasetSize} reveals a clear trend where the average relative error in drag coefficient predictions decreases as the percentage of the dataset used for training increases. This trend is consistent for both datasets, underscoring the common machine learning principle that more training data generally leads to better model performance.
The improved performance of the DrivAerNet Dataset across all sizes of training data highlights the critical role of bigger datasets in machine learning models for aerodynamics and further establishes the value of the DrivAerNet dataset, which is significantly larger than previous open-source datasets.

The figure also indicates that RegDGCNN yields better performance on the DrivAerNet dataset compared to the dataset from ShapeNet. This can be attributed to several factors: 
\begin{itemize}
    \item The large variation in shapes within ShapeNet does not correspond with an adequate number of samples to encompass the entire range of aerodynamic drag values.
    \item ShapeNet cars were modeled as single-bodied entities, omitting crucial details like wheels and underbodies that are vital for accurate aerodynamic modeling.
    \item There is a considerable variation in mesh resolutions across the ShapeNet dataset, potentially leading to inconsistencies in aerodynamic predictions.
\end{itemize}

This analysis serves to demonstrate the generalization capabilities of our model, emphasizing the objective of developing models that effectively generalize to out-of-distribution designs.

\section{Limitations and Future Work}
\label{sec:Limitations}
This section discusses the limitations of our study. Despite careful selection to ensure a balance between detail and computational efficiency, the model's parameterization faces an inherent limitation. This stems from the trade-off between the compactness of the representation and the flexibility needed to capture a broad range of aerodynamic phenomena. Consequently, while our approach offers valuable insights for many applications, it may not fully encompass all aerodynamic variabilities pertinent to automotive engineering.
The dataset comprises 4,000 instances, which, while significant, might not fully capture the broad spectrum of real-world automotive designs. A feature importance analysis should be conducted on the parametric data to efficiently determine the impact of various parameters on aerodynamic performance. This approach will facilitate more efficient sampling and guide the prioritization of parameters in future dataset expansions.  While our dataset is large and of high-fidelity, it is important to acknowledge that we are still in the early stages of approaching the scale and foundational impact seen in AI fields like image processing and natural language processing, where large datasets are a norm. 

One of the key challenges in applying graph-based methods like RegDGCNN is the significant GPU memory requirements. This is due to the need to compute all pairwise distances between points, which can be highly memory-intensive. Moreover, the non-uniform density of point clouds introduces additional complexities; a fixed $k$-nearest neighbors approach may not be suitable for areas with varying densities of nodes. Another limitation is that, in its current form, RegDGCNN does not reduce the number of points during the forward pass for large-scale point clouds, leading to high computational demands and potentially limiting the model's scalability to even larger datasets. Additionally, our focus was primarily on drag prediction; however, we plan to extend the application of RegDGCNN to include surface field predictions in future work.
Addressing these challenges will be crucial for advancing the capabilities and applications of graph-based neural networks in processing complex aerodynamic data.

\section{Conclusion}
\label{sec:Conclusion}

In our conclusion, we highlight the distinct advantages of DrivAerNet, which, by focusing on detailed geometric modifications, outperforms broader datasets such as those referenced in~\cite{song2023surrogate, li2023geometryinformed, Rios2021}, especially in the context of real-world aerodynamic design applications. Additionally, the compact RegDGCNN model, with 3 million parameters and a 10MB size, efficiently estimates drag in just 1.2 seconds for industry-standard designs with 540k mesh faces, significantly outpacing traditional CFD simulations. Moreover, our RegDGCNN model, in particular, showcases superior performance by directly processing 3D meshes, thereby eliminating the need for 2D image rendering or the generation of Signed Distance Functions (SDF), which simplifies the preprocessing stages and increases the model's accessibility. Importantly, the RegDGCNN model's ability to deliver precise drag predictions without requiring water-tight meshes highlights its adaptability and effectiveness in leveraging real-world data. Through the expansion of the DrivAerNet dataset from 560 to 2800 samples, we achieved a remarkable reduction in error by approximately 75$\%$.  Similarly, on the dataset from \cite{song2023surrogate}, increasing the training samples from 1270 to 6352 led to a 56$\%$ decrease in error, underscoring the substantial impact of dataset scale on enhancing the performance of deep learning models in aerodynamic studies. The DrivAerNet dataset, which is generated through the morphing of a baseline parametric model to produce high-quality industry-standard designs, proves to be a better approach compared to the designs found in ShapeNet. This emphasizes the critical role of large, detailed, high-fidelity datasets in crafting models capable of adeptly handling the complexities inherent in aerodynamic surrogate modeling.

\section{Dataset Name}

\textbf{Name:} \textit{DrivAerNet}

\section{Motivation}
\subsection{Why was the dataset created?}
\textit{The DrivAerNet dataset was developed to provide a benchmark for training deep learning models in aerodynamic car design. It was created to address the lack of publicly available datasets that include industry-standard parametric car geometries and high-fidelity computational fluid dynamics (CFD) simulations, enabling researchers to develop and test machine learning models for engineering applications.}

\subsection{Who created the dataset?}
\textit{DrivAerNet was created by the Design Computation and Digital Engineering (DeCoDE) Lab at the Massachusetts Institute of Technology (MIT) in collaboration with the 3D AI Lab at the Technical University of Munich (TUM).}

\subsection{Who funded the creation of the dataset?}
\textit{The dataset creation was primarily supported by computational resources provided by MIT. Mohamed Elrefaie was supported by MIT DeCoDE lab with additional support from TUM and the DAAD IFI Fellowship.}

\section{Composition}
\subsection{What do the instances that comprise the dataset represent?}
\textit{Each instance in DrivAerNet represents a parametric 3D car geometry along with high-fidelity CFD simulation results. The dataset includes 3D fields for velocity, pressure, and wall-shear stresses, as well as aerodynamic coefficients for various car configurations.}

\subsection{How many instances are there in total?}
\textit{DrivAerNet consists of 4,000 full-domain 3D CFD simulations.}

\subsection{What data does each instance consist of?}
\textit{Each instance contains car geometry files in STL format, CFD simulation results in VTK format, aerodynamic coefficients, and tabular parametric design data. Additional annotations and point cloud representations are also included for machine learning applications.}

\subsection{Is there any missing data? If so, how much?}
\textit{No.}

\subsection{How is the data associated with each instance organized?}
\textit{Each design is stored in a structured folder system, with separate files for geometry and simulation results ensuring easy access and reproducibility.}

\section{Collection Process}
\subsection{How was the data collected?}
\textit{The dataset was generated using parametric car designs that were simulated in OpenFOAM, with mesh generation performed using SnappyHexMesh. Python scripts automated the data generation and extraction process.}

\subsection{Who was involved in the data collection process?}
\textit{The dataset was developed by researchers from MIT and TUM, utilizing automated pipelines to generate, process, and validate the simulations.}

\subsection{Over what timeframe was the data collected?}
\textit{The dataset was created over the course of 2023 and 2024.}

\subsection{How was it decided which data to collect and which to exclude?}
\textit{Only geometries that passed validation checks for CFD simulations were included, ensuring accurate and meaningful aerodynamic analysis.}

\section{Preprocessing, Cleaning, and Labeling}
\subsection{Was any preprocessing, cleaning, or labeling done?}
\textit{Yes, preprocessing was applied to remove geometries with non-manifold edges or intersecting surfaces. The dataset also includes labeled aerodynamic coefficients and part annotations for machine learning applications.}

\subsection{How was this done and by whom?}
\textit{The preprocessing was automated using Python scripts developed by the MIT and TUM research teams.}

\section{Uses}
\subsection{For what tasks in Engineering Design is the dataset suitable?}
\textit{DrivAerNet is suitable for aerodynamic shape optimization, turbulence modeling, generative AI for car design, reduced-order modeling, and surrogate modeling of aerodynamic forces.}

\subsection{Has the dataset been used for any tasks already?}
\textit{Yes, DrivAerNet has been utilized for aerodynamic surrogate modeling of the aerodynamic drag.}

\subsection{What performance metrics are relevant for assessing tasks using this dataset?}
\textit{Relevant metrics include drag and lift coefficient prediction accuracy, R-squared for regression models, and Chamfer distance for design similarity tasks.}

\section{Distribution}
\subsection{How is the dataset distributed?}
\textit{DrivAerNet is publicly available on GitHub~\url{https://github.com/Mohamedelrefaie/DrivAerNet} and will also be hosted on the Harvard Dataverse~\url{https://dataverse.harvard.edu/dataverse/DrivAerNet} for long-term availability.}

\subsection{Are there any restrictions or licenses on its use?}
\textit{DrivAerNet is released under the Creative Commons Attribution-NonCommercial 4.0 International License (CC BY-NC 4.0) and is exclusively for non-commercial research and educational purposes.}

\subsection{What is the duration of availability for the dataset as provided by the authors?}
\textit{The dataset will remain publicly available indefinitely, with updates provided as needed.}

\section{Maintenance}
\subsection{Who is responsible for dataset maintenance?}
\textit{The dataset is maintained by the Harvard Dataverse team and the authors.}

\subsection{How can individuals submit corrections or updates?}
\textit{Corrections and updates can be submitted via GitHub issues~\url{https://github.com/Mohamedelrefaie/DrivAerNet/issues} and pull requests~\url{https://github.com/Mohamedelrefaie/DrivAerNet/pulls}.}

\subsection{Is there a versioning system in place?}
\textit{Yes, the dataset follows a versioning system to track updates and modifications, ensuring reproducibility.}

\section*{Nomenclature}
\small
\section*{Roman and Greek Letters}
\begin{tabular}{@{}p{0.15\columnwidth} p{0.05\columnwidth} p{0.7\columnwidth}@{}}
$u_1, u_2, u_3$ & = & Velocity components in the $x_1$, $x_2$, and $x_3$ directions, respectively (m/s) \\
$u_\infty$ & = & freestream velocity (m/s) \\
$p$ & = & Pressure (Pa) \\
$\rho$ & = & Fluid density (kg/m$^3$) \\
$\mu$ & = & Dynamic viscosity (Pa·s or kg/(m·s)) \\
$f_1, f_2, f_3$ & = & Body force components in the $x_1$, $x_2$, and $x_3$ directions (N) \\
$k$ & = & Turbulence kinetic energy (m$^2$/s$^2$) \\
$\omega$ & = & Specific dissipation rate (1/s) \\
$G$ & = & Generation of turbulence kinetic energy due to mean velocity gradients \\
$D_k, D_\omega$ & = & Diffusivities of turbulence kinetic energy and specific dissipation rate, respectively \\
$S_k, S_\omega$ & = & Source terms for turbulence kinetic energy and specific dissipation rate, respectively \\
$\mathbf{S}$ & = & Magnitude of the mean rate-of-strain tensor \\
$\nu$ & = & Kinematic viscosity, $\mu / \rho$ (m$^2$/s) \\
$\nu_t$ & = & Turbulence viscosity (m$^2$/s) \\
$\gamma, \beta, \beta^*$ & = & Model constants \\
$\tau_{ij}$ & = & Reynolds stress tensor (Pa) \\
\end{tabular}

\vspace{0.5em}
\section*{Dimensionless Groups}
\begin{tabular}{@{}p{0.15\columnwidth} p{0.05\columnwidth} p{0.7\columnwidth}@{}}
Re & = & Reynolds number, $\rho u_\infty l / \mu$ \\
$C_d$ & = & Drag coefficient \\
$C_l$ & = & Lift coefficient \\
$C_{l_r}$ & = & Rear lift coefficient \\
$C_{l_f}$ & = & Front lift coefficient \\
$C_p$ & = & Pressure coefficient, $\frac{p - p_{\infty}}{0.5 \rho u_{\infty}^2}$ \\
\end{tabular}

\vspace{0.5em}
\section*{Machine Learning and Graph Terms}
\begin{tabular}{@{}p{0.15\columnwidth} p{0.05\columnwidth} p{0.7\columnwidth}@{}}
$G$ & = & Graph representing nodes and edges \\
$X$ & = & Node features of the graph \\
$h_{ij}$ & = & Edge features between node $i$ and node $j$ \\
$\Theta$ & = & Parameters of the EdgeConv layers \\
$\phi$ & = & Parameters of the Fully Connected layers \\
$\mathcal{N}(i)$ & = & Neighborhood of node $i$ \\
$\max$ & = & Max pooling operation for global feature aggregation \\
\end{tabular}

\vspace{0.5em}
\section*{Chamfer Distance Terms}
\begin{tabular}{@{}p{0.15\columnwidth} p{0.05\columnwidth} p{0.7\columnwidth}@{}}
$CD(A, B)$ & = & Chamfer distance between point sets $A$ and $B$ \\
$|A|, |B|$ & = & Cardinality (number of points) in point sets $A$ and $B$, respectively \\
$a, b$ & = & Points in point sets $A$ and $B$, respectively \\
$\|a - b\|^2$ & = & Squared Euclidean distance between point $a$ and point $b$ \\
\end{tabular}

\vspace{0.5em}
\section*{Evaluation Metrics}
\begin{tabular}{@{}p{0.15\columnwidth} p{0.05\columnwidth} p{0.7\columnwidth}@{}}
$C_d$ & = & Actual drag coefficient obtained from CFD simulations \\
$\hat{C}_d$ & = & Predicted drag coefficient from the surrogate model \\
$\bar{C}_d$ & = & Mean drag coefficient \\
$n$ & = & Number of samples \\
MSE & = & Mean Squared Error, $\frac{1}{n} \sum_{i=1}^{n} (C_{d_{i}} - \hat{C}_{d_{i}})^2$ \\
$R^2$ & = & Coefficient of Determination, $1 - \frac{\sum_{i=1}^{n} (C_{d_{i}} - \hat{C}_{d_{i}})^2}{\sum_{i=1}^{n} (C_{d_{i}} - \bar{C}_{d})^2}$ \\
\end{tabular}

\vspace{0.5em}
\section*{Subscripts and Superscripts}
\begin{tabular}{@{}p{0.15\columnwidth} p{0.05\columnwidth} p{0.7\columnwidth}@{}}
$i, j, k$ & = & Indices used in Einstein summation notation, representing Cartesian coordinates \\
$\langle \cdot \rangle$ & = & Denotes Reynolds-averaged quantities \\
$\prime$ & = & Denotes fluctuating quantities \\
\end{tabular}

\vspace{0.5em}
\section*{Miscellaneous} 
\begin{tabular}{@{}p{0.15\columnwidth} p{0.05\columnwidth} p{0.7\columnwidth}@{}}
$F_{23}$ & = & Blending function in the SST model \\
$a_1, b_1$ & = & Model constants used in the SST model \\
\end{tabular}

\vspace{0.5em}
\section*{Acronyms}
\begin{tabular}{@{}p{0.15\columnwidth} p{0.05\columnwidth} p{0.7\columnwidth}@{}}
    SDF & = & Signed Distance Fields \\
    CFD & = & Computational Fluid Dynamics \\
    HPC & = & High-Performance Computing \\
    RegDGCNN & = & Dynamic Graph Convolutional Neural Network for Regression \\
    MLP & = & Multi-Layer Perceptron \\
    FC & = & Fully Connected \\
    RANS & = & Reynolds-Averaged Navier-Stokes \\ 
    URANS & = & Unsteady Reynolds-Averaged Navier-Stokes \\
    LBM & = & Lattice Boltzmann Method \\
    LES & = & Large Eddy Simulation \\
    SST & = & Shear Stress Transport \\
    PDEs & = & Partial Differential Equations \\ 
    DNN & = & Deep Neural Networks \\
    PCA & = & Principal Component Analysis \\
    GINO & = & Geometry-Informed Neural Operator \\
    GeoCA & = & Geometry-Guided Conditional Adaptation \\
    KNN & = & K-Nearest Neighbors \\
    OpenFOAM & = & Open Source Field Operation and Manipulation \\
    SHM & = & SnappyHexMesh \\
    STL & = & Stereolithography \\
    VTK & = & Visualization Toolkit \\
    CSV & = & Comma-Separated Values \\
    KDE & = & Kernel Density Estimation \\
    ESE & = & Enhanced Stochastic Evolutionary \\
\end{tabular}

\normalsize

\section{Acknowledgements}

The research work of Mohamed Elrefaie was supported by the German Academic Exchange Service (DAAD) under the IFI - "Internationale Forschungsaufenthalte für Informatikerinnen und Informatiker" Grant and the Department of Mechanical Engineering at MIT. The authors thank Florin Morar from BETA CAE Systems USA Inc for his support with morphing the DrivAer model in ANSA®. We extend our gratitude to Christian Breitsamter for his role in facilitating the organizational aspects of this research.

\footnotesize
\bibliographystyle{abbrv}
\bibliography{refs.bib}

\end{document}